\documentclass[acmtog,nonacm]{acmart}

\usepackage{epigraph}
\usepackage{nth}
\usepackage{algorithm}
\usepackage{algpseudocode}

\usepackage{booktabs}
\usepackage{graphicx}
\usepackage{soul}
\usepackage{dsfont}
\usepackage{gensymb}
\usepackage{microtype}
\usepackage{pifont}
\usepackage{threeparttable}
\usepackage{collect}
\usepackage{xspace}
\usepackage{xfrac}
\usepackage[capitalise]{cleveref}
\usepackage{natbib}
\usepackage{array}
\usepackage{multirow}
\usepackage{wrapfig}
\usepackage{url}
\usepackage[nolist]{acronym}
\usepackage{siunitx}

\usepackage{pifont}

\creflabelformat{equation}{#2\textup{#1}#3}

\DeclareGraphicsExtensions{.png,.jpg,.pdf,.ai,.psd}
\def\figurePath{figures/}

\acmJournal{TOG}

\acmSubmissionID{321}

\usepackage{xr}
\makeatletter
\newcommand*{\addFileDependency}[1]{
  \typeout{(#1)}
  \@addtofilelist{#1}
  \IfFileExists{#1}{}{\typeout{No file #1.}}
}
\makeatother

\newcommand{\refFig}[1]{Fig.~\ref{fig:#1}}

\newcommand{\refTab}[1]{Tab.~\ref{tab:#1}}

\newcommand{\refSec}[1]{Sec.~\ref{sec:#1}}

\newcommand{\refEq}[1]{Eq.~\ref{eq:#1}}

\newcommand{\refAlg}[1]{Alg.~\ref{alg:#1}}

\usepackage[bold=0.55]{xfakebold}
\newcommand{\winner}[1]{\setBold #1\unsetBold}

\usepackage{icomma}

\newcommand{\myfigure}[2]{%
    \begin{figure}[htb]%
    \centering\includegraphics*[width = \linewidth]{\figurePath#1}%
    \caption{#2}%
    \label{fig:#1}%
    \end{figure}%
}

\newcommand{\mycfigure}[2]{%
    \begin{figure*}[htb]%
    \centering\includegraphics*[width = \linewidth]{\figurePath#1}%
    \caption{#2}%
    \label{fig:#1}%
    \end{figure*}%
}

\soulregister\ref7
\soulregister\cite7
\soulregister\refFig7
\soulregister\refAlg7
\soulregister\cite7
\soulregister\ref7
\soulregister\pageref7
\soulregister\shortcite7
\soulregister\eg0
\soulregister\ie0
\soulregister\etal0

\DeclareGraphicsExtensions{.png,.jpg,.pdf,.ai,.psd}
\newcommand{\mysection}[2]{\section{#1}\label{sec:#2}}
\newcommand{\mysubsection}[2]{\subsection{#1}\label{sec:#2}}
\newcommand{\mysubsubsection}[2]{\subsubsection{#1}\label{sec:#2}}

\definecolor{colorA}{HTML}{4285f4}
\definecolor{colorB}{HTML}{ea4335}
\definecolor{colorC}{HTML}{fbbc04}
\definecolor{colorD}{HTML}{34a853}
\definecolor{colorE}{HTML}{ff6d01}
\definecolor{colorF}{HTML}{46bdc6}
\definecolor{colorG}{HTML}{000000}
\definecolor{colorH}{HTML}{777777}
\definecolor{colorI}{HTML}{bdd6ff}
\definecolor{colorJ}{HTML}{6a9e6f}

\newcommand{\numOfData}{22\xspace}
\newcommand{\numOfFlashData}{21\xspace}
\newcommand{\numOfRealData}{14\xspace}
\newcommand{\numOfSynthData}{7\xspace}
\newcommand{\material}[1]{\textsc{#1}}

\begin{acronym}
\acro{NN}{neural network}
\acro{MLP}{multi-layer perceptron}
\acro{BRDF}{bi-directional reflectance distribution function}
\acro{svBRDF}{spatially-varying \ac{BRDF}}
\acro{ODE}{ordinary differential equation}
\acro{FOV}{field of view}

\acro{AD}{automatic differentiation}
\acro{MLP}{multi-layer perceptron}
\acro{NCA}{neural celluar automata}
\acro{DyNCA}{dynamic \ac{NCA}}
\acro{ODE}{ordinary differential equation}
\acro{PDE}{partial differential equation}
\acro{SDE}{stocastic differential equation}
\acro{CNN}{convolutional neural network}
\acro{IVP}{initial value problem}
\acro{GN}[\texttt{GN}]{group normalization}
\acro{AdaGN}[\texttt{AdaGN}]{adaptive \acl{GN}}
\acro{SWD}{sliced Wasserstein distance}
\acro{NFE}{network forward execution}
\end{acronym}
\usepackage{bm}

\newcommand{\eg}{e.g.,\ }
\newcommand{\ie}{i.e.,\ }
\newcommand{\etal}{et~al.\ }

\newcommand{\mymath}[2]{
    \newcommand{#1}{\TextOrMath{$#2$\xspace}{#2}}
}

\mymath{\latentState}{\mathbf z}
\mymath{\timeCoord}{t}
\mymath{\parameters}{\theta}
\mymath{\appearance}{\mathbf{L}}
\mymath{\targetAppearance}{\hat{\appearance}}
\mymath{\ode}{\mathbf f}
\mymath{\noise}{\xi}
\mymath{\timeWarmUp}{t_{\mathrm{I}}}
\mymath{\timeStart}{t_{\mathrm{S}}}
\mymath{\timeEnd}{t_{\mathrm{E}}}
\mymath{\timeSolve}{t}
\mymath{\timeNow}{t_\mathrm{N}}
\mymath{\project}{\rho}
\mymath{\projectMatrix}{\mathbf{E}}
\mymath{\intd}{\mathrm d}
\mymath{\stats}{\mathtt{VGG}}
\mymath{\distance}{\mathcal D}
\mymath{\loss}{\mathcal L}

\mymath{\render}{\mathcal R}
\mymath{\viewingDir}{\bm{\omega_\textbf{o}}}
\mymath{\lightingDir}{\bm{\omega_\text{i}}}

\mymath{\spaceCoord}{\mathbf x}
\mymath{\odex}{\mathrm f}
\mymath{\gaussian}{\mathcal{N}}
\mymath{\uniform}{\mathcal{U}}
\mymath{\channels}{C}
\mymath{\augChannel}{9}
\mymath{\stateRange}{\mathbb{R}}
\mymath{\imageChannels}{3}
\mymath{\pixelRange}{\mathbb{R}}
\mymath{\timeRange}{\mathbb{R}}
\mymath{\height}{H}
\mymath{\width}{W}
\mymath{\timeEmbeddingDim}{L}
\mymath{\hiddenLayerCount}{N}
\mymath{\timeEncoder}{\gamma}
\mymath{\numOfFrames}{M}

\mymath{\refreshRate}{R}
\mymath{\numOfIter}{N}

\mymath{\gold}{\text{gold}}
\mymath{\steel}{\text{steel}}
\mymath{\goldFrame}{\appearance_{\gold}}
\mymath{\encodedFrame}{\appearance_{\text{enc}}}
\mymath{\steelFrame}{\appearance_{\steel}}
\mymath{\mean}{\mu}
\mymath{\std}{\sigma}

\mymath{\randomCropSymbol}{\mathtt{C}}
\mymath{\randomCropIdx}{k}
\mymath{\randomCrop}{\randomCropSymbol_{\randomCropIdx}}

\mymath{\randomShuffleSymbol}{\mathtt{S}}
\mymath{\randomShuffleIdx}{k}
\mymath{\randomShuffle}{\randomShuffleSymbol_{\randomShuffleIdx}}

\mymath{\weight}{w}
\mymath{\weightGlobal}{\weight_{\mathrm{G}}}
\mymath{\weightLocal}{\weight_{\mathrm{L}}}
\mymath{\weightInit}{\weight_{\mathrm{I}}}
\mymath{\distanceGlobal}{\distance_\mathrm{Global}}
\mymath{\distanceLocal}{\distance_\mathrm{Local}}
\mymath{\distanceInit}{\distance_\mathrm{Init}}

\mymath{\firstOrder}{1^{\text{st}}}
\mymath{\zeroOrder}{0^{\text{th}}}

\newcommand{\refSecSupplImplementation}{Suppl.~Sec.~2}

\begin{document}

\acmSubmissionID{281}
\setcopyright{acmlicensed}
\acmJournal{TOG}
\acmYear{2024} \acmVolume{43} \acmNumber{6} \acmArticle{256} \acmMonth{12}\acmDOI{10.1145/3687900}

\renewcommand{\eg}{\textit{e.g.}, }
\renewcommand{\ie}{\textit{i.e.}, }
\newcommand{\wrt}{w.r.t.}
\newcommand{\etc}{etc.}

\ccsdesc[500]{Computing methodologies~Computer graphics}
\ccsdesc[500]{Computing methodologies~Reflectance modeling}
\ccsdesc[500]{Computing methodologies~Machine learning}

\keywords{Material Appearance; Dynamic Texture; Dynamic (sv)BRDF; Neural Differential Equation; Video}

\title{Neural Differential Appearance Equations}

\author{Chen Liu}
\affiliation{%
	\institution{University College London}
	\country{United Kingdom}
}
\email{chen.liu.21@ucl.ac.uk}

\author{Tobias Ritschel}
\affiliation{%
	\institution{University College London}
	\country{United Kingdom}
}
\email{t.ritschel@ucl.ac.uk}

\begin{teaserfigure}  
    \includegraphics[width=\textwidth]{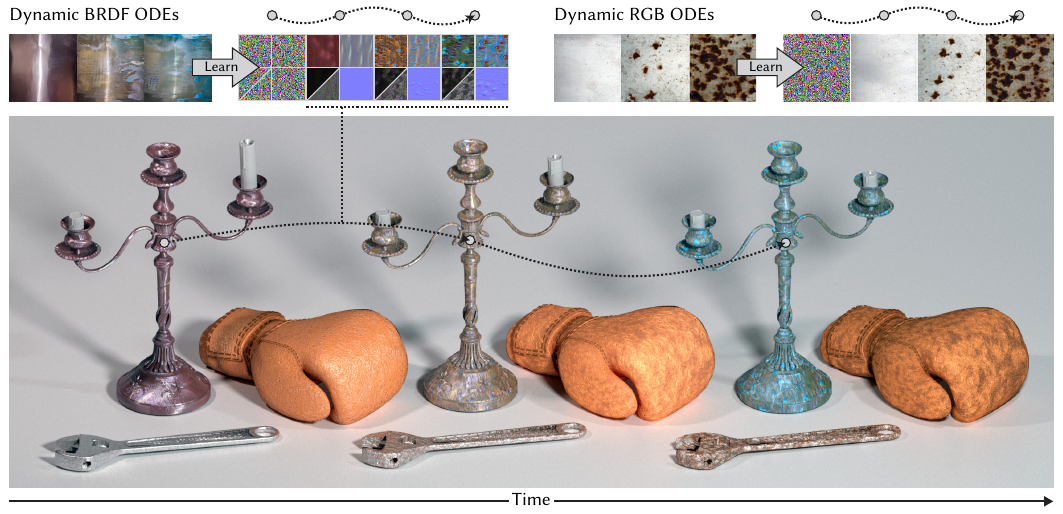}
    \caption{
    We propose to learn a model of dynamic visual appearance using \acsp{ODE} (dashed lines) from image sequences. 
    An example of dynamic \acs{svBRDF} maps and an example of dynamic RGB textures are shown at the top. 
    We learn \acsp{ODE} to synthesize them from a sequence of flash-lit images or an RGB texture video, respectively.
    The rendering at the bottom makes use of our neural BRDF ODEs of patinating copper, rusting iron, and weathering leather.
    }
    \label{fig:Teaser}
\end{teaserfigure}

\begin{abstract}
We propose a method to reproduce dynamic appearance textures with space-stationary but time-varying visual statistics.
While most previous work decomposes dynamic textures into static appearance and motion, we focus on dynamic appearance that results not from motion but variations of fundamental properties, such as rusting, decaying, melting, and weathering.
To this end, we adopt the neural \ac{ODE} to learn the underlying dynamics of appearance from a target exemplar.
We simulate the \ac{ODE} in two phases.
At the ``warm-up'' phase, the \ac{ODE} diffuses a random noise to an initial state.
We then constrain the further evolution of this \ac{ODE} to replicate the evolution of visual feature statistics in the exemplar during the generation phase.
The particular innovation of this work is the neural \ac{ODE} achieving both denoising and evolution for dynamics synthesis, with a proposed temporal training scheme.
We study both relightable (\acs{BRDF}) and non-relightable (RGB) appearance models.
For both we introduce new pilot datasets, allowing, for the first time, to study such phenomena:
For RGB we provide \numOfData dynamic textures acquired from free online sources;
For \acsp{BRDF}, we further acquire a dataset of \numOfFlashData flash-lit videos of time-varying materials, enabled by a simple-to-construct setup.
Our experiments show that our method consistently yields realistic and coherent results, whereas prior works falter under pronounced temporal appearance variations.
A user study confirms our approach is preferred to previous work for such exemplars.
\end{abstract}

\maketitle

\setlength\epigraphwidth{5.8cm}
\epigraph{One side holding that the ship remained the same, and the other contending that it was not the same.}{Plutarch, \textit{Life of Theseus 23.1}}

\mysection{Introduction}{Introduction}
Formation of patterns is ubiquitous in nature, where they are rarely static but usually evolve over time, \eg color changing for deciduous leaves.
The pioneering work of \citet{turing1952chemical} describes the formation of biological patterns in terms of differential equations to model reaction and diffusion of chemical concentration.
In graphics, manual design of such equations was used to synthesize textures \cite{witkin1991reaction,turk1991generating}.
Unfortunately, no approach exists to automatically discover them from data.
This is the aim of the present paper.

Dynamic textures have been explored for decades \cite{peteri2010dyntex,hadji2018new,tesfaldet2018two,pajouheshgar2023dynca} under the assumption that textures can be factored into static appearance and motion, excluding numerous natural phenomena mentioned above and illustrated in \refFig{Teaser}.
While at every point in time, the image is a texture, \ie it has uniform statistics over space, these statistics change over time.
In particular, the development of rust spots into rust is not motion but a complete change of shape, color, creation of new elements, and deletion of others.

Popular approaches in exemplar-based dynamic texture synthesis \cite{funke2017synthesising,tesfaldet2018two,zhang2021dynamic,xie2017synthesizing,pajouheshgar2023dynca} usually extend the Gram loss \cite{gatys2015texture} with motion, resulting in what is referred to as ``two-stream'' (appearance and motion) methods. 
These existing methods only consider dynamic textures that are spatio-temporally homogeneous in appearance and motion. 
The focus of this paper is to solve temporally heterogeneous appearance.
We demonstrate that the reproduction of time-varying, but not time-stationary appearance is non-trivial and simply imposing a per-frame loss during pixel-wise optimization does not suffice to recover consistency between frames, even with motion matched.
Instead, we learn a neural \ac{ODE} that simulates the dynamics of latent states and projects them to dynamic colors. 
In practice, we introduce a training algorithm with online iterations and refreshes for dynamic appearance \acp{ODE}.

While the visual appearance of a material might undergo drastic variation due to external influences as well as changes in illumination, we might say the material remains the same.
Capturing and reproducing the effect for both of those factors is challenging and finding light and time-invariant representations might contribute towards understanding a material’s essence.
To this end, we generalize RGB dynamic appearance to relightable \ac{BRDF} maps, with a differentiable renderer and flash images \cite{aittala2016reflectance, henzler2021generative}.

To evaluate our method, we collect two new datasets.
The first consists of \numOfData representative dynamic textures with time-varying statistics, mostly curated from free online sources. 
The second consists of \numOfFlashData exemplars captured under flashlight illumination using a lightweight setup,  allowing to compute relightable appearance.

We evaluate our method in terms of realism and temporal coherence, both qualitatively and quantitatively, outperforming other comparative methods.
It includes a user study that compares our method favorably to baselines in human perception.

In summary, our contributions are:
\begin{itemize}
    \item A method to discover differential equations to model time-varying appearance from image observations;
    \item A dataset of space-stationary, time-varying RGB textures;
    \item Combination of this model with inverse rendering to become relightable;
    \item A dataset of flash-lit space-stationary, time-varying textures.
\end{itemize}

Our code and dataset will be released at \url{https://github.com/ryushinn/ode-appearance}.

\mysection{Previous Work}{PreviousWork}

\mysubsection{Dynamic Texture Synthesis}{Dynamic Texture Synthesis}

Dynamic textures have been the focus of extensive research. 
Early works \cite{doretto2003dynamic,costantini2008higher} propose a linear dynamical system to analyze and synthesize dynamic textures.
The far-reaching Gram loss of \citet{gatys2015texture} shapes modern approaches to (dynamic) texture synthesis. 
They discover that features in intermediate layers of a pre-trained VGG network for object classification are powerful textural descriptors.
So, pixel-wise optimizing an image to have the same Gram matrices of VGG features as the exemplar generates a novel texture. 
A convolutional network can also be trained to map noises to textures using Gram loss \cite{ulyanov2017improved}.
\citet{funke2017synthesising} generate dynamic textures with concatenated VGG features in a temporal window.
Extending \citet{gatys2015texture}, the Two-stream method by \citet{tesfaldet2018two} incorporates features from another pre-trained optical flow network as temporal motion statistics, which disentangles appearance and motion components of a dynamic texture.
\citet{zhang2021dynamic} modify the Two-stream method with a shifted Gram loss and a frame sampling strategy to add long-range correlation for space and time.
\citet{pajouheshgar2023dynca} enable real-time synthesis by training \ac{NCA} for dynamic textures.

However, one commonality for not only current methods but also existing datasets \cite{peteri2010dyntex,hadji2018new} of dynamic textures is that the dynamics have been mainly explained as motion. 
Thus, target exemplars are mostly animated static images, such as wavy sea, swinging flags, and flames, whose appearance is temporally stationary. 
Changes of appearance --let alone reflectance-- are rarely considered.
In the broader field of dynamic video synthesis, the same constraint remains \cite{nam2019end, zhang2020dtvnet, nikankin2023sinfusion}, and a sufficiently large dataset is further required to train generative models \cite{brooks2022generating}. 
SinGAN-GIF and SinFusion \cite{nikankin2023sinfusion, arora2021singan} can be trained on a single video but present limited temporal changes.
In contrast, given a single texture video, our method can synthesize realistic and consistent time-varying appearance without relying on any motion features.
As real-world appearances often change dynamically, \eg rusting, numerous applications could benefit from our method that extends to such phenomena. 

\mysubsection{Time-varying Materials}{Time-varying Reflectance}

\paragraph{BRDFs/Materials}
An impressive body of work exists to acquire materials, ranging from dedicated setups
\cite{marschner1998inverse,matusik2003data}, optimization-based methods \cite{lombardi2015reflectance, aittala2016reflectance, guo2020materialgan}, to deep learning methods \cite{georgoulis2017reflectance,deschaintre2018single, henzler2021generative, sartor2023matfusion, zhou2022look}.
We here do not try to generally solve the inversion task, but to over-fit to a single material \cite{aittala2015two,fischer2022metappearance}, albeit it changes over time.

Closest to ours in aim is the approach of \citet{gu2006time} and \citet{sun2007time}.
They use an existing BDRF acquisition infrastructure to take many \ac{BRDF} samples over time.
It is then assumed that there is a dynamic change of reflectance occurring at any pixel that is low-dimensional.
Besides avoiding the heavy capture setup, our approach does not make this assumption and can deal with drastic, deforming, and non-linear changes of reflectance.
Their approach however supports arbitrary \ac{BRDF}, while we assume a parametric \ac{BRDF} model to be sufficient, as well as stationarity.
Finally, ours is a generative model both in spatial arrangement and temporal dynamics.

\paragraph{Dynamics}
The change of appearance has been approached from the angle of weathering \cite{dorsey2005modeling,chen2005visual,xuey2008image,merillou2008survey}, patina \cite{dorsey2006modeling}, dusting \cite{hsu1995simulating}, corrosion \cite{merillou2001corrosion}, cracking paint \cite{paquette2002simulation}, organic decay \cite{kider2011fruit}, flow of substances \cite{dorsey1996flow} and plant growing \cite{desbenoit2004simulating}.
Devising --and sometimes also using-- these models requires expert chemical, physical, or engineering knowledge in the particular material's domain \cite{callisterjr2007materials}.
Our approach discovers a latent space that might be similar to the underlying chemical-physical processes and states but is not exposed to the user or prescribed by the developer.

Appearance manifolds \cite{wang2006appearance} discover how a certain spatial appearance pattern can change into another one over time.
Our approach has a few similarities as it discovers a change that defines a manifold locally.
While their exemplars can be handled as RGB curves for a single instance, it has not been demonstrated for generation and/or \acp{BRDF}.

\paragraph{Diffusion and automata}
From another angle, generic methods such as reaction-diffusion 
\cite{turing1952chemical,witkin1991reaction,turk1991generating} or cellular automata \cite{gobron2001crack} are more similar to our approach, but the rules they use are typically not discovered from data but need to be set up by an expert user.

A recent work that inspired our approach is the idea of revisiting reaction-diffusion or automata from the perspective of deep learning. \citet{niklasson2021self} developed neural automata that first were used only to fit a single image.
We will revisit the background of \ac{NCA} in \refSec{Neural Differential Equation}.
This was later extended to dynamics \cite{pajouheshgar2023mesh} but under the assumption that dynamics are deformations or motion flows.
We extended this idea to fully discover the transition rules of some latent chemical or physical process that manifests as dynamic RGB textures with time-varying visual statistics, based on which we consider relighting and reflectance.

\mysubsection{Neural Differential Equation}{Neural Differential Equation}

The long belief that differential equations are pertinent to dynamic textures comes from Turing's work \cite{turing1952chemical}, which claims that patterning is governed by the diffusion and reaction process of some chemicals associated with pigment production. 
This idea is then employed to synthesize natural textures \cite{witkin1991reaction}, \eg giraffe dots, zebra stripes, and reptile skin, by designing and solving the differential equations of the reaction-diffusion process. 

Recently, research on neural diffusion processes \cite{sohldickstein2015deep,ho2020denoising,song2019generative,rombach2022high, heitz2023iterative, lipman2023flow} has exploded and began learning (stochastic) differential equations through data.
Modern diffusion-based models achieve state-of-the-art for various image synthesis tasks by matching the distribution of a large dataset of natural images.
Our \ac{ODE}, in contrast, is not trained on a large dataset but overfits each exemplar.
Our approach is also different from SinFusion \cite{nikankin2023sinfusion}, as it is not based on a score function.
We instead simulate the equation to produce training signals \cite{chen2019neural}.
Moreover, it does not use a pre-learned latent space \cite{rombach2022high} but finds a space for every exemplar it overfits to.
Additionally, the paramount distinction is that our \ac{ODE} is simulated to generate not only a single image (denoising in the ``warm-up'' phase) but a sequence of images to form a video (evolving in the generation phase) from noise, as detailed in \refSec{Objective}.

One more relevant work to ours is \ac{NCA}.
\ac{NCA} is a special differential equation proposed by \citet{mordvintsev2020growing} for general purposes \cite{mordvintsev2020thread}. 
Specifically, a grid of cells gets iteratively updated due to their interactions with neighbors. 
The update rule represented by neural networks is applied to each cell for each iteration. 
\citet{niklasson2021self} explore the application of \ac{NCA} in static texture synthesis.
Furthermore, \ac{DyNCA} \cite{pajouheshgar2023dynca} extends \cite{niklasson2021self} to control the motion in dynamic textures using a pre-trained motion network. 
However, synthesizing videos with significant temporal changes of appearance using differential equations has not been sufficiently investigated, which is what we focus on here.

\mysection{Our Approach}{OurApproach}
\mycfigure{Overview}{Overview of our approach.
Input is a time series of images shown in the first row.
Output is a model that enables sampling new texture instances shown by the pink curve in the bottom row.
This is achieved by defining an \ac{ODE} in a higher-dimensional latent space, shown as the blue curve.
We learn an \ac{ODE} vector field (orange arrows) guiding the update of the latent state.
We start from noise, and after a warm-up phase (no exemplar supervision, shown in the first row), the new instance evolves.
We project the latent coordinates to RGB space.
To make the texture relightable, we involve two projections: one from latent to BRDF parameters (yellow curve), and then a projection that is rendering itself, conditioned on light and view.
}

We will next elaborate on our approach.
The explanation will be on general "appearance" (which can mean simple RGB, or more complicated, relightable \ac{svBRDF}) and only highlight differences.
Input and output (\refSec{InputOutput}) are the same, so is the dynamic latent space, as well as the overall optimization objective (\refSec{latent_appearance}, and \refSec{Objective}).
The main difference is in the projection from latent to appearance (\refSec{Projection}).

\mysubsection{Input and Output}{InputOutput}
Our input is a sequence of RGB images, presenting a dynamic appearance where the statistics of visual features change over time.
One exemplar is shown at the top of \refFig{Overview}.
We denote this sequence as $
\targetAppearance(\timeCoord):
\mathbb \timeRange
\rightarrow
\mathbb \pixelRange^{3\times\height\times\width}
$
where \height and \width are image resolution, and $\timeCoord\in[\timeStart, \timeEnd]$ is the temporal coordinate of the sequence in a time interval between a start \timeStart and an end \timeEnd.
The output is a time-varying appearance model $\appearance_{\parameters}(\timeCoord):
\mathbb \timeRange
\rightarrow
\mathbb \pixelRange^{3\times\height\times\width}
$ such that at any time $\timeCoord$ in the interval $[\timeStart, \timeEnd]$ it matches the target $\targetAppearance(\timeCoord)$.
We will learn different parameters \parameters from each exemplar.

We model the dynamics $\appearance_{\parameters}$ with neural \acp{ODE} \cite{chen2019neural}. 
Working out dynamics in direct RGB appearance space seems a daunting task, as noted by \citet{dupont2019augmented}, in particular, if we wish to make the next step to also relight the result.
So instead of RGB, we combine established knowledge of the physics of rendering with a model of the unknown (chemical or biological) physics of material evaluation we want to discover.
To this end, we will work in a space of latent coordinates, but for the case of relighting, some internal states are useful and interpretable as common BRDF parameters used in legacy rendering systems.

\mysubsection{Latent Space}{latent_appearance}

We represent the cause of appearance as a continuous sequence of latent states $
\latentState( \timeCoord):
\mathbb \timeRange
\rightarrow
\mathbb \stateRange^{\channels\times\height\times\width}$. 
The latent states are of shape $\channels\times\height\times\width$, where 
\channels is the dimension of states, and \height and \width are the spatial extent of the grid.

We do not make assumptions about those latent states, which can be chemical concentrations, velocities, or gradients with respect to space or time.
They are latent, in the sense that learning will use them for things useful to solve the task but we do not prescribe what they are or require expert knowledge to define them as was required by previous expert systems \cite{witkin1991reaction}.

Now, we formulate the temporal progression of latents as a learnable neural \ac{ODE} parameterized by $\parameters$, which can be later projected to the temporal progression of appearance:
\begin{equation}
    \frac{\intd \latentState}{\intd \timeCoord}(\timeCoord) = \ode_{\parameters}(\latentState(\timeCoord), \timeCoord).
\end{equation}
We model the neural \ac{ODE} as a convolutional neural network $
\ode_{\parameters}(\latentState, \timeCoord): 
(\stateRange^{\channels\times\height\times\width}, \timeRange)
\rightarrow 
\stateRange^{\channels\times\height\times\width}
$ predicts the state update given the current state and time.
Details on the architecture and its training are found in \refSec{network_details} and \refSec{training_details}.

Using this \ac{ODE}, we can solve $\latentState$ to the future latent state at any time $\timeNow$ from an initial condition $\latentState(\timeWarmUp)$, with an appropriate \ac{ODE} solver as in:
\begin{equation}
    \latentState_{\parameters}(\timeNow) = \latentState(\timeWarmUp) + \int_{\timeWarmUp}^{\timeNow}\ode_{\parameters}(\latentState(\timeSolve), \timeSolve) \intd \timeSolve.
\end{equation}

\mysubsection{Projection}{Projection}
Latent states \latentState relate to visual appearance, which is our supervision, in terms of a projection $\project:\stateRange^{\channels\times\height\times\width}\rightarrow\pixelRange^{3\times\height\times\width}$ to RGB colors as in $\appearance_{\parameters}(\timeCoord) = \project(\latentState_{\parameters}( \timeCoord))$.
Projection could be thought of, \eg to define what colors a certain combination of chemicals in a decaying leaf would have.
As the projection is fixed, but the latent structure is learned, optimization will adapt the latent states so that they result in an RGB color that matches the supervision, once projected.
We use channel augmentation proposed in prior works \cite{mordvintsev2020growing, mordvintsev2021texture}.
Projection is done differently for the RGB and the relightable variant of our method:

\paragraph{RGB variant}
For an instance of RGB textures, our latent space comprises a total of $\channels = 3+9$ channels and the projection \project simply involves extracting the first $3$ channels:
\begin{equation}
    \project^{\text{RGB}}(\latentState) = \projectMatrix\latentState,
\end{equation}
where $\projectMatrix$ is an extraction matrix with a top-left identity sub-matrix and zeros elsewhere.
The choice of nine augmented states is a manual one that seems to work well with a single value used for all exemplars in this paper, but could be tuned.

\paragraph{Relightable / svBRDF variant}
We augment in the same way for \ac{svBRDF} \acp{ODE}, with one distinction, that the appearance states are not directly RGB values but \ac{BRDF} parameters. 
The latent states have $\channels = 9 + 9$ channels with the first 9 channels of \ac{svBRDF} parameters and 9 augmented channels.
Therefore we couple the extraction with a forward differentiable renderer, as detailed in \refSec{renderer}, to project.

Specifically, the projection for \ac{svBRDF} \acp{ODE} is:
\begin{equation}
    \label{eq:brdf_project}
    \project^{\ac{svBRDF}}(\latentState)
    =\render(\projectMatrix\latentState, \lightingDir, \viewingDir),
\end{equation}
where $\render$ denotes a (differentiable) renderer and \projectMatrix still is the extraction matrix, that now maps from latent space to BRDF parameter space.
The renderer takes a set of \ac{svBRDF} maps and shades it under the given light-view condition $(\lightingDir, \viewingDir)$, where $\lightingDir$ is the lighting map that specifies the lighting direction for each pixel and $\viewingDir$ is the same viewing map.

\mysubsection{Objective}{Objective}
Matching the time-varying target appearance with parameters \parameters means that for every possible, normally-distributed initial latent state $\latentState(\timeWarmUp)\sim\gaussian$, the solution of the ODE has to match the target during the interval $[\timeStart, \timeEnd]$:
\begin{equation}
\label{eq:goal}
\min_{\parameters} \quad
\mathbb E_{\latentState(\timeWarmUp)\sim\gaussian,\timeNow\sim{\uniform(\timeStart, \timeEnd)}}
\left[
    \distance(
        \appearance_{\parameters}(\timeNow), \ 
        \targetAppearance(\timeNow)
    )
\right]
,
\end{equation}
where \distance is the textural distance between two images.
The implementation of the loss is detailed in \refSec{loss_details}. For now it can be assumed a differentiable black box that compares the perception of textures.
Stationarity \cite{aittala2015two}, \ie the same statistics across space, is the key to untangling the latents and BRDF in space and time.

Although we only care about latent states $\latentState_{\parameters}(\timeCoord)$ within \timeStart and \timeEnd, a key innovation of this work is to solve an \ac{ODE} not from \timeStart to \timeEnd but to start at an earlier time \timeWarmUp for ``warm-up'', so $\timeWarmUp < \timeStart < \timeEnd$. 
Our approach has two phases (\refFig{Overview}):
The warm-up phase ($\timeCoord\in[\timeWarmUp,\timeStart]$) begins with random noise, \ie $\latentState(\timeWarmUp)\sim\gaussian$, and the noise gradually evolves to form an appearance at \timeStart from which the generation phase ($\timeCoord\in[\timeStart, \timeEnd]$) can reproduce the dynamic appearance.
While the appearance is updated by the \ac{ODE} to match the changing $\targetAppearance(\timeCoord)$ between \timeStart and \timeEnd, the only supervision on what the appearance needs to be between \timeWarmUp and \timeStart is the first frame of the target $\targetAppearance(\timeStart)$. 
Therefore the only reason why the appearance changes during the ``warm-up'' is to ``become ready'' to correctly evolve in the deployed interval, \ie \timeStart to \timeEnd.
It is also noted at the top of \refFig{Overview} that no supervision is present during this interval $[\timeWarmUp, \timeStart]$, where the noises learn their own way to become diverse and non-repeatable textures or \acp{svBRDF} via the training signal of merely one image frame.

Warm-up plays an important conceptual role in our approach, ensuring the \ac{ODE} can first drive the noise to an initial appearance and hence marking a clear start for synthesized videos.
Without it, the first video frame would simply be noise.
The generation phase after warm-up uses the neural ODE to produce subsequent appearance evolution, which, to our knowledge, is used in generative modeling tasks for the first time. 
In inference, diverse dynamic appearances can be synthesized by solving the trained ODE from a random initial state. 
Notably, in our approach we learn a single ODE model to govern both phases, handling not only denoising but also evolution.
Although we did not conduct an ablation study without the warm-up phase, we trust it is evident that the absence of this phase would render the method ineffective for useful dynamic appearance synthesis.

\mysection{Implementation}{Implementation}

To make the above ideas work, several further innovations need to be made, which we discuss in this section.
More implementation details can be found in \refSecSupplImplementation.

\mysubsection{Training}{training_details}
Effective training relies on using the right \ac{ODE} modeling and efficient estimation of the loss expectation, all to be explained next.

\myfigure{network}{An overview of our ODE UNet. The dashed lines are skip-concatenations. The Down blocks first halve the image size and then double the number of channels. The Up blocks do the opposite. For activation, We use Swish by default, except the second-to-last $1 \times 1$ convolution, which employs Sigmoid for boundedness.}

\paragraph{ODE modeling}
Our neural ODE $\ode_\parameters$ is a UNet \cite{ronneberger2015u} inspired by modern diffusion models \cite{dhariwal2021diffusion}, as shown in \refFig{network}.
Our UNet mainly consists of the \texttt{ConvBlock}, a $3\times 3$ convolution followed by an \ac{AdaGN} and a self-attention layer (\texttt{Self-Attn}).
We apply circular padding to all convolutional layers, which enables the seamless synthesis of tileable RGB textures and \ac{svBRDF} maps.
We further discuss the \ac{ODE} architecture concerning our task of dynamic appearance modeling in \refSec{network_details}.
This network is similar to the denoiser used in diffusion models as it also drives an \ac{ODE}, but without relying on score functions for training.

As our \ac{ODE} solver, we employ the adaptive Heun method, setting a tolerance of $10^{-2}$ for both training and inference. 
This configuration provides the optimal trade-off between efficiency and accuracy in our experiments. 

\paragraph{Estimation}
A direct way to optimize \refEq{goal} is to, for each iteration, simulate \ac{ODE} from a random noise at $\timeWarmUp$ to a random $\timeNow\in[\timeStart, \timeEnd]$, compute the loss between $\targetAppearance(\timeNow)$ and $\appearance_{\parameters}(\timeNow)$ and update the model. 
This scheme is compute-intensive, as the simulation has to start from scratch and solves for an average integral time of $\frac{\timeStart + \timeEnd}{2} - \timeWarmUp$ in every iteration, covering more than half of the entire integral range.
We propose an effective and efficient training scheme for dynamic appearance \acp{ODE} in \refAlg{training}. 

\begin{algorithm}
\caption{Dynamic appearance \ac{ODE} training.}
\label{alg:training}
\begin{algorithmic}[1]
\Require
ODE parameters \parameters, 
number of iterations \numOfIter, 
refresh rate \refreshRate,
target dynamic appearance \targetAppearance.
\Ensure the trained \ac{ODE} parameters \parameters
\For{$i=0, 1, ..., \numOfIter$}
\State $k=i \bmod \refreshRate$
\If{$k == 0$}
    \State $\latentState_0 = \Call{Sample}{\gaussian}$
    \State $\timeCoord_0, \timeCoord_1 = \timeWarmUp, \timeStart$
\Else
    \State $\timeCoord_1$ =\Call{Sample}{ $\uniform[\frac{\timeEnd-\timeStart}{\refreshRate - 1} \cdot (k-1), \frac{\timeEnd-\timeStart}{\refreshRate - 1} \cdot k]$} 
\EndIf
\State $\latentState_1$=
\Call{ODESolve}
{$\latentState_0, \timeCoord_0, \timeCoord_1; \ode_{\parameters}$}

\State \parameters = 
\Call{Update}{$\mathrm d\distance(
\targetAppearance(\timeCoord_1), \project(\latentState_1))/\mathrm d\parameters$}

\State $\latentState_0, \timeCoord_0 = \latentState_1, \timeCoord_1$
\EndFor
\end{algorithmic}
\end{algorithm}

We estimate the expectation of \refEq{goal} in an ``online'' manner: the integral range is divided into small segments and approximated progressively.
For every iteration, we solve from $\latentState_0$ at $\timeCoord_0$ to $\latentState_1$ at $\timeCoord_1$ and update \parameters using $\latentState_1$; 
In the next iteration, we will reuse the last state $(\latentState_1, \timeCoord_1)$ as the new starting point $(\latentState_0, \timeCoord_0)$ to solve from, instead of starting from scratch, and sample a new supervision point $\timeCoord_1$.
At the first iteration, we begin with $\latentState_0$ sampled as Gaussian noise at $\timeCoord_0=\timeWarmUp$ and supervise at $\timeCoord_1=\timeStart$.
For several following iterations, we progressively solve the \ac{ODE}, with the supervision points $\timeCoord_1$ uniformly sampled on stratified ranges covering the entire generation phase.
Once we have iterated over all ranges, the above procedure will be repeated for a new cycle. 

We refresh and repeat the process at a specific rate $\refreshRate$.
Online iterations significantly decrease the average integral time to $\frac{\timeEnd-\timeStart}{\refreshRate - 1}$ by trading the accuracy of gradients, as the true $\latentState(\timeCoord_0)$ is now approximated using the last state.
A shorter integral time for each iteration will make more approximations to solve the entire domain with errors accumulating.
We tune $\refreshRate = 6$ for all exemplars, which experimentally works well.

Online iterations are actually efficient in both time and memory with smaller integral measures. 
Adjoint gradient proposed by \citet{chen2019neural} is also memory-efficient although we opt to not use it, as discussed in \refSec{adjoint_discussion}.

Moreover, $\timeStart$ is importance-sampled every \refreshRate iteration with refreshes.
Consequently, the first frame of appearance provides more supervision than others.
It is similar to adopting a larger training weight for the first frame and consistently improves synthesis quality.
We discuss the effect of this in \refSec{steps_discussion}.
Prior \ac{NCA} works \cite{mordvintsev2021texture,pajouheshgar2023dynca} likewise utilize a mechanism named ``checkpointing with restarts'' to prevent the network from forgetting how to transform noises to textures.

\mysubsection{Differentiable Rendering}{renderer}
The key to our BRDF approach is to chain the dynamic appearance idea with differentiable rendering and the aim is modeling the dynamics of \ac{BRDF} parameters given observations. 

\paragraph{BRDF model}
The model employed is the Cook-Torrance microfacet model \shortcite{cook1982reflectance} with anisotropic GGX normal distribution \cite{heitz2014understanding}, Smith-GGX geometry term \shortcite{smith1967geometrical}, and Schlick's Fresnel approximation \shortcite{schlick1994inexpensive}. 
We also complement this microfacet model with an extra diffuse Lambertian lobe, which is not physically based but works well in an uncalibrated context.
Therefore, our BRDF parameterization is the RGB diffuse albedo for the diffuse lobe, and the RGB specular albedo and the roughness in two axes for the Cook-Torrance model. 
Including a normal map, parameterized by a one-channel height map, our \ac{svBRDF} maps have a total of nine channels.

Some materials we study have effects this shading model cannot reproduce, such as subsurface scattering.
As ``all models are wrong'' \cite{box1976science}, this is no drawback, as long as our model explains data (\ie dynamics of appearance) better than other models: melting cheese without scattering is still better than melting cheese without melting. 

\paragraph{Shading geometry}
We assume the captured sample is parallel to the camera plane, as the real physical setup in \refSec{ProtocolBRDF}.
Thus, the two direction maps \lightingDir and \viewingDir can be derived from the shading geometry inferred by the camera \ac{FOV}.
We use a \ac{FOV} of 50 degrees, the \ac{FOV} of our capture device, and assume a unit-length distance between the camera and the sample in the renderer. 
It means that the virtual geometry is now subject to an unknown global scaling factor relative to the physical setup. 
Although the global geometric scaling does not affect the scene being rendered with correct angles, the light intensity attenuates by the distance. 
Previous works \cite{aittala2016reflectance, henzler2021generative, guo2020materialgan} either assume a fixed intensity or try to estimate it from the brightness of images. 
This will force the \ac{BRDF} parameters to compensate for the discrepancy in the real irradiance received and thereby is more likely to induce the baked lighting artifacts.
Instead, we simply add the intensity as an additional parameter to be optimized.

\paragraph{Light and camera intrinsics}
The \emph{exact} replication of the physical setup within the forward renderer is impossible without expertise.
For example, various devices apply distinct tone-mapping operations while taking photographs;
moreover, the true attenuation of lighting depends on the specification of the flashlight. 
These intrinsics are mostly proprietary and not disclosed to users.
Calibrating these details contradicts our goal of ensuring a lightweight and accessible acquisition process for non-experts. Consequently, we opt for common approximations for them, such as the inverse power gamma correction for tone-mapping and the inverse square falloff for attenuation.

\mysubsection{Loss}{loss_details}
The loss is implemented using three components: 
\begin{equation}
\distance = 
\weightGlobal\distanceGlobal+
\weightLocal\distanceLocal+
\weightInit\distanceInit,
\end{equation}
which are global and local visual statistics, and an additional initialization loss.

\paragraph{Global Stats}
To compare the global visual statistics between the \targetAppearance and \appearance images, we first transform them into a perceptual descriptor space, the commonly used VGG feature space \cite{gatys2015texture}, and then measure the distance between the two feature distributions using the \ac{SWD} \cite{heitz2021sliced}:
\begin{equation}
    \label{eq:SWD}
    \distanceGlobal(\appearance, \targetAppearance) = \mathtt{SWD}(\stats(\appearance), \stats(\targetAppearance)).
\end{equation}

\paragraph{Local Stats}
After shading, relightable / svBRDF results cannot be analyzed by global statistics, but only by local statistics of patches under varying light-view directions, where stationarity implicitly exists \cite{aittala2015two,henzler2021generative,zhou2023photomat}:
\begin{equation}
    \label{eq:brdf_distance}
    \distanceLocal(\appearance, \targetAppearance) = 
    \sum_{\randomCropIdx}\distanceGlobal(
        \randomCrop(\appearance),
        \randomCrop(\targetAppearance)
    ).
\end{equation}
\randomCrop denotes random cropping.
For efficiency, we only simulate \acp{ODE} at the size of the cropped region in every iteration.

\paragraph{Init}
It is also well-established in the literature that an appropriate initialization is crucial for plausible \ac{svBRDF} estimation from photographs when ground-truth supervision is unavailable \cite{aittala2016reflectance, li2017modeling, gao2019deep, guo2020materialgan}.
The existing methods such as using an off-the-shelf estimation network, however, are not applicable to \ac{ODE} weight initialization.
We propose a novel method which optimizes the \ac{ODE} using a special loss for weight initialization:
\begin{equation}
    \label{eq:init_distance}
    \distanceInit(\appearance, \targetAppearance) = 
    \sum_{\randomShuffleIdx}
    \left (
        \randomShuffle(\appearance) - 
        \randomShuffle(\targetAppearance)
    \right )^2,
\end{equation}
where \randomShuffle denotes spatially random shuffling of the original map and $\randomShuffle(\appearance):=\render(\projectMatrix\latentState, \randomShuffle(\lightingDir), \randomShuffle(\viewingDir))$.
This loss resembles the random crop loss in \refEq{brdf_distance}, and can be considered as one-pixel random crops which preserve no spatial structure.
It actually imposes an extreme of stationarity, forcing that each pixel in the synthesized \ac{svBRDF} maps, when rendered under all observed light-view angles, must match the corresponding pixel of the target at that angle.
It initializes \acp{ODE} to produce spatially constant \ac{svBRDF} maps for each time point, which describe an ``average'' of the target flash image.

Practically, only one loss term is turned on for each training phase, \ie one of weights is 1 while others are 0. For dynamic BRDF appearance, we initially train \acp{ODE} with $\weightInit=1$ for sufficient iterations to establish a robust initialization, from which we continue training with $\weightLocal=1$.
For dynamic RGB appearance, we set $\weightGlobal=1$.

\mysection{Results}{results}
We here discuss our main results, first for non-relightable RGB (\refSec{rgb_results}) and second for its relightable counterpart (\refSec{brdf_results}), which also produces svBRDFs that can be used in legacy, simulation-based renderers.
The section concludes with a use study, to compare our results to other methods (\refSec{user_study}).

We recommend visiting our supplemental for animations and hence more comprehensive appreciation of all our results.

\mysubsection{Dynamic RGB Texture Application}{rgb_results}

Our results are based on data captured using a certain protocol (\refSec{protocol_rgb}), compared to baselines by some metrics which lead to both qualitative (\refSec{qualitative_rgb}) and quantitative (\refSec{quantitative_rgb}) results.

\mysubsubsection{Protocol}{protocol_rgb}

\paragraph{Data collection}
Dynamic textures considered in this paper are not well represented in current datasets \cite{peteri2010dyntex,hadji2018new}. 
We collect our own \numOfData textures from the Internet (\eg YouTube), migrate and modify from existing datasets, and capture some of them in the wild.
Our data comprehensively spans a spectrum of processes, capturing the intricate dynamics of chemical reactions (\eg rusting steel), biological growth (\eg sprouting seeds), physical transformations (\eg crystallization), and various combinations of the three; see the row ``exemplar'' in \refFig{DynamicTextures}.

\paragraph{Data Processing}
Exemplars are casually calibrated and captured, which sometimes causes flashes, jitters, and shifts.
We process videos as inputs by space-cropping at the center and resizing them into $128^2$ resolution and time-scaling them to five-second videos, each with 100 frames. 
We match the \ac{ODE} time scale to the actual length of dynamics by setting $\timeWarmUp=-1$, $\timeStart=0$, and $\timeEnd=5$, \ie the duration of the input. 
We train the \ac{ODE} with noises in the same size as the input while we can generate novel dynamic textures in an arbitrary size in inference. 

\paragraph{Baselines}
We compare to five baselines in dynamic texture/video synthesis, proposed by (i) \citet{gatys2015texture}, (ii) \citet{ulyanov2017improved}, (iii) \citet{tesfaldet2018two} (Two-stream), (iv) \citet{pajouheshgar2023dynca} (\ac{DyNCA}), and (v) \citet{nikankin2023sinfusion} (SinFusion).
We synthesize a video with 100 frames using each method.

The first two methods are designed for only one image of texture as the input.
We apply them to our input video frame-wise to generate results.

Given a target video of dynamic textures, the Two-stream method can natively synthesize videos, but with all frames sharing the same visual statistics.
For fair comparisons, we \emph{upgrade} Two-stream method with our idea of temporal appearance loss, \ie frame-wise optimizing the appearance loss while optimizing the proposed motion loss between frames.

Moreover, for fairness and during inference of all baselines, we start from the same initial noise for every frame to further improve temporal coherence of all baselines.

\paragraph{Metrics}
We measure the quality of synthesized dynamic textures in two dimensions: realism and temporal coherence. 

We evaluate \emph{realism} by computing the average textural loss, such as Gram loss and \ac{SWD} loss, between the 100 pairs of synthesized frames and corresponding target frames.

The metrics for video texture \emph{coherence} are, however, still under active research.
Our coherence metric is based on the recent research \cite{henaff2019perceptual, harrington2022exploring} of perceptual straightness, where the authors find that a more naturally coherent video is perceived by humans as a straighter trajectory in the space of perceptual responses.
Specifically, given a video, we transform all frames to the VGG feature space and then estimate the average curvature of this feature path as the ``non-straightness''.
Smaller average curvatures indicate a more linear and straighter trajectory, hence a more coherent input video.

\mysubsubsection{Qualitative Results}{qualitative_rgb}

\mycfigure{DynamicTextures}{Results of our method on six exemplars (six blocks).
In each block, time goes left to right, and on the top we show a frame of the exemplar, and below our re-synthesis at that time point.
The re-synthesis is performed on a random field three times as high as the original image, to demonstrate we can produce infinite, diverse non-repeating samples.
The change of leaf color from green to red starts with medium-sized and dark-red islands that grow in size followed by a global sweep into a darker brown.
Our method reproduces well in colors and sizes with the same global transformation evolving in the form of a spatial ``front''.
The growing sprouts leave little to desire, given the complexity of the deformation and the specificity of the shapes involved.
Copper successfully crystallizes with crystal shapes and colors matched.
The mold textures are accurate over the bread, with dark specks and white hyphae developing.
We do well in reproducing the initial combs of melting honey and their partial destruction, even including some letting fluid-like deformation besides the appearance change to darker beige.
The dehydrating radish undergoes large changes which we capture well, from colors to patterns, even shadows.
}

We present six textures sampled by our method in \refFig{DynamicTextures}, and in the caption, we qualitatively analyze their realistic appearance in detail, while the coherent shape transformations are also evident. 

\mycfigure{MainComparisons_1}{Comparisons between our method and baselines (three exemplars).
Our method successfully reproduces the evolution of target appearances: The melon rind turns from fresh orange to dark brown with black spots appearing;
Water droplets on the window freeze into ice filaments and eventually form frost patterns;
The dynamics of cracks are quite accurate, starting from the periphery and gradually progressing.
}

We then demonstrate comparative results from our method, the upgraded Two-stream, \ac{DyNCA}, and SinFusion in \refFig{MainComparisons_1}. 
Similar to ours, the Two-stream method can generate realistic textures for each frame; 
However, it fails to maintain consistency across frames, with anomalies such as cracks shifting from the top-right to the bottom-left corner. 
It also introduces artifacts in an attempt to align with the optical flow motion which these three exemplars do not involve.
\ac{DyNCA}, which also relies on a motion module, encounters the same issue and only animates static textures with ill-defined motions without capturing the true dynamics.
SinFusion can only simulate a short segment of the dynamics, resulting in limited temporal variation.

\mysubsubsection{Quantitative Results}{quantitative_rgb}

\begin{table}
\caption{
Numerical results for all RGB and svBRDF generation methods.
}
\label{tab:main_numerical_results}

\resizebox{\columnwidth}{!}{%
\begin{tabular}{rrccccc}
\toprule
 &  & \multicolumn{2}{c}{Gram $\downarrow$} & \multicolumn{2}{c}{SWD $\downarrow$} & Non-Str. $\downarrow$ \\ 
 \midrule
\multirow{7}{*}{\rotatebox[origin=c]{90}{RGB}} & Gatys & \multicolumn{2}{c}{\winner{0.005}} & \multicolumn{2}{c}{\winner{0.013}} & 2.050 \\
 & Ulyanov & \multicolumn{2}{c}{0.032} & \multicolumn{2}{c}{0.043} & 2.073 \\
 & Two-stream & \multicolumn{2}{c}{0.006} & \multicolumn{2}{c}{0.015} & 1.777 \\
 & DyNCA & \multicolumn{2}{c}{3.216} & \multicolumn{2}{c}{1.924} & 1.499 \\
 & SinFusion & \multicolumn{2}{c}{4.372} & \multicolumn{2}{c}{1.900} & 1.749 \\
 & Ours & \multicolumn{2}{c}{0.042} & \multicolumn{2}{c}{0.050} & \winner{0.720} \\
 & Exemplar & \multicolumn{4}{c}{} & 1.531 \\ 
 \midrule
 &  & \small{Center} & \small{Novel} & \small{Center} & \small{Novel} &  \\ 
 \cmidrule(lr){3-4}
 \cmidrule(lr){5-6}
\multirow{7}{*}{\rotatebox[origin=c]{90}{\ac{svBRDF}}} & MaterialGAN & 0.101 & 0.092 & 0.128 & 0.161 & 2.059 \\
 & Henzler & 0.170 & 0.148 & 0.197 & 0.233 & 2.086 \\
 & MatFusion & 0.053 & 0.049 & 0.049 & 0.066 & 1.588 \\
 & Look-ahead & 0.049 & 0.085 & 0.036 & 0.075 & 1.678 \\
 & Deschaintre & 0.374 & 0.226 & 0.444 & 0.298 & 1.784 \\
 & Ours & \winner{0.020} & \winner{0.020} & \winner{0.034} & \winner{0.038} & \winner{0.984} \\
 & Exemplar & \multicolumn{4}{c}{} & 1.390 \\ 
 \bottomrule
\end{tabular}

}
\end{table}

The quantitative results for all methods are in \refTab{main_numerical_results}.
The \ac{DyNCA} and SinFusion struggle to produce plausible results for these dynamic textures characterized by significant temporal appearance changes.
As anticipated, frame-wise optimization methods excel in realism but lack coherence, a factor of critical perceptual importance for synthesized videos.
In contrast, our method consistently yields the most linear trajectories.
Furthermore, our user study, detailed in \refSec{user_study}, confirms that our method generates samples with realism that is indistinguishable from that achieved by frame-wise optimization methods.
We also evaluate the coherence of the target dynamic textures.
An intriguing observation is that our \acp{ODE} produce videos with greater consistency than the exemplars themselves. 
This is due to the inherent evolutionary properties of \acp{ODE}, which effectively smooth out the capturing artifacts such as flashes, shifts, and jitters present in the target input.

\mysubsection{Dynamic BRDF Texture Application}{brdf_results}

\mysubsubsection{Protocol}{ProtocolBRDF}

Our acquisition aims to capture flash images of a time-variant process.

\paragraph{Setup}
The samples we consider are of similar and moderate spatial extend, \qty{10}{\centi\metre\squared}, limited height of less than \qty1{\centi\meter}.
Different from previous works \cite{aittala2015two, aittala2016reflectance, henzler2021generative} that shot only one or two images during a short period (\ie a few seconds), capturing up to hours or even days is sensitive to the environmental lighting change. 
The actual setup consists of a mobile phone, mounted to a stand, and a flashlight equipped by the phone.
Our setup is placed in an optic enclosure to ensure the flash lighting dominates all the time.
An illustration is shown in \refFig{Setup}.

We capture raw images for each material at appropriate regular intervals to record the temporal appearance variations throughout the process, akin to creating a time-lapse video. 
While capturing, we position the camera at an appropriate distance to exclusively frame the material sample and fix all exposure settings and the focal length for every image taken to ensure consistency.
Our assumptions are the same as for \citet{aittala2016reflectance} or \citet{henzler2021generative}.

\myfigure{Setup}{Schematic and photo of our acquisition setup. The optic enclosure is omitted in the photo for simplicity.}

\mycfigure{Dataset}{Start \textbf{(top)} and end frames \textbf{(bottom)} for a subset of the time-varying material appearance we captured.}

\paragraph{Data}
With the capture setup explained, we collect $\numOfRealData$ real-world phenomena of dynamic materials.
We showcase some samples from the dataset in \refFig{Dataset}.

The data for each material consists of approximately 500 raw images, each with a resolution of $4034 \times 3024$.
In addition, we also build $\numOfSynthData$ synthetic scenes in Blender \cite{blender2024} complying with that setup, in order to comprehensively evaluate our method using the ground truths of relighting.
To our best knowledge, this compilation provides the first-ever dataset of time-varying material appearances, capturing various temporally varying properties such as surface bumps, metallicity, wetness, and temperature.

We sample 100 frames from each raw time-lapse flash-lit video, crop them to square at the center, and resize to $256^2$. 
We empirically set $\timeWarmUp=-2$, $\timeStart=0$, and $\timeEnd=10$ for \ac{svBRDF} \acp{ODE}.
During training, we perform one-forth random crops so we solve \acp{ODE} in $128^2$ resolution.
While our trained \acp{ODE} support infinite-size synthesis, we sample $256^2$ noises to generate results for comparisons here.

\paragraph{Baselines}
To the best of our knowledge, we are the first approach dedicated to re-synthesizing dynamic \ac{BRDF} textures from image sequences.
Other methods either are not dynamic \cite{aittala2016reflectance,henzler2021generative} or require a BRDF scanner \cite{gu2006time,sun2007time}.
We compare to five recent works in \ac{svBRDF} estimation by: (i) \citet{guo2020materialgan} (MaterialGAN), (ii) \citet{henzler2021generative}, (iii) \citet{sartor2023matfusion} (MatFusion), (iv) \citet{deschaintre2018single}, and (v) \citet{zhou2022look} (Look-ahead).
All these methods aim to estimate a set of static \ac{svBRDF} maps from one image or multiple images of the same material, and are applied to each frame of our data to generate the final dynamic output.
Same as in \refSec{protocol_rgb}, we fairly fix the noise to synthesize each frame more consistently for generative methods \cite{henzler2021generative, sartor2023matfusion}.

\paragraph{Metrics}
Previous methods \cite{sartor2023matfusion, zhou2022look, deschaintre2018single} typically assess outcomes using ground truth \ac{svBRDF} maps. 
However, such maps are unavailable for real-world captures, like our dataset. 
Additionally, synthetic \ac{svBRDF} maps heavily depend on the specific \ac{BRDF} model and the renderer used, and sometimes represent different properties even with the same name. 
It involves intricate conversions and alignments to adapt \ac{svBRDF} maps used in different works.

Therefore, we opt to focus on the rendered images, treating the synthesized \ac{svBRDF} maps, the \ac{BRDF} parameterization, and the renderer jointly as the internal component of each method. 
We measure realism and coherence (same metrics as \refSec{protocol_rgb}) solely based on the rendered results.
We show the \ac{svBRDF} maps only for qualitative analysis.

To evaluate each method, we relight the generated dynamic \ac{svBRDF} maps under the flash lighting at the center, identical to the capture setup, and three novel flash lightings at the top, left, and bottom-right, under which the true quality of the \ac{svBRDF} maps can be effectively exhibited. 
We also render the synthetic scenes with this setting to get the ground truth relit renderings.
Realism metrics are measured by comparing center-lit renderings with the input exemplars for all data, and novel-lit renderings with ground truth relightings for synthetic data.
Coherence is assessed for all videos.

\mysubsubsection{Qualitative Results}{QualitativeBRDF}

\mycfigure{DynamicSVBRDF}{
Dynamic \ac{svBRDF} synthesized by our \acp{ODE} for six exemplars.
The two blocks in the middle column are synthetic data, and the rest is our real captured data.
In each block, the time progresses from left to right.
From top to bottom are the exemplar, renderings by center flash lighting, relighting by a novel flashlight in the left, and the generated \ac{svBRDF} maps, respectively. 
The \ac{svBRDF} maps consist of a \textbf{D}iffuse map, a \textbf{S}pecular map, two \textbf{R}oughness maps (\textbf{u} for the horizontal axis and \textbf{v} for the vertical axis), and a \textbf{N}ormal map.
Their layout is marked in the first inset at the top-left block.
}

In \refFig{Teaser} we demonstrate a 3D scene where objects are textured with our generated \ac{svBRDF} maps. We further showcase six dynamic \ac{svBRDF} results of our method in \refFig{DynamicSVBRDF} and will detail their appearance and dynamics from left to right, and top to bottom.

The first block is \material{Clay Solidifying}, which we capture very well.
As the dots in clay bumps are solidifying and changing to strips over time, we can observe that highlight scattering becomes more uniform.
Some cyan colors appear on the two albedo maps, reflecting a variation process of reflectance from wet clay to dry clay.

\material{Metal Rusing} in the second block is also a successful example.
Although parsing the evolution of these \ac{BRDF} maps might be difficult for such an intricate process, they are combined, when rendered, to well explain the dynamic appearance of a rusting metal surface. 
The relighting result is also accurate which confirms that we generate a set \ac{svBRDF} maps without baked lighting artifacts.

The material of \material{Honeycomb Melting} is challenging, with layered reflectance and varying shapes due to comb destruction.
Our result reproduces them surprisingly well.
In the first frame, the comb shapes are evident as polygonal bumps, which gradually disappear in the following frames and become flat.
The albedo and roughness maps also show some comb patterns and produce plausible rendered appearances, especially for the middle frame where the appearance of honeycomb and melted beeswax are well mixed.

The fourth material, \material{Copper Patinating}, changes the color from the uniform dark pink of raw copper to the varied rusty green of the patina.
The material exhibits strong anisotropic highlights in the early frames which our anisotropic BRDF model successfully captures as two distinctly different roughness maps: low horizontal roughness and high vertical roughness.
This gives rise to the elongated highlight shape, which moves to the left as anticipated when relit with a left flashlight.
This exemplar is less spatially stationary with patina and rust growing in some specific regions. 
Our result, in contrast, stationarizes appearance across the image.

We synthesize a realistic sample of the overall dynamics of \material{Ceramics Dirt-covering}. 
The surface roughness gradually increases as observed from the more and more diffuse appearance, so as the albedo map becomes dark.
However, the diagonal stripe pattern is not faithfully reconstructed and appears somewhat distorted.
This issue arises from the textural loss we adopt, which compares visual statistics without strictly preserving the structure.

Finally, the last block shows \material{Cheese Melting}.
While being an exotic material, which might not really fit our model assumptions due to some sub-surface scattering, the new appearance is quite plausible.
Most importantly, bumps indeed recovered are generated as bumps, not as dark albedo, as we see from the relighting image.
The bumps also change shape from something like stripes to isolated dots to waves.
As expected, the albedo tint goes from white to brown.
We note the expected correlation between different aspects of the material, such as normals and white albedo. 

In \refFig{Comparisons_SVBRDF}, we qualitatively compare results from MaterialGAN, \citeauthor{henzler2021generative}, and MatFusion.
MatFusion has realistic results close to ours, while instead of generating new samples it is estimating the original material, \eg the identical rust pattern as the input.
The results of \citet{henzler2021generative} are diverse but, when rendered, show an appearance that is overly diffuse without correct highlights and is wrongly dark on the other side.
The baked lighting is the most pronounced in MaterialGAN, which results in noticeable artifacts when relit by novel light conditions.

\mysubsubsection{Quantitative Results}{QuantitativeBRDF}
The overall numerical results are reported in \refTab{main_numerical_results}.
Our method demonstrates superior realism compared to the baselines. 
Our similar performance for both center and novel lighting conditions indicates that the synthesized \ac{svBRDF} maps effectively capture the underlying material properties. 
While MatFusion and Look-ahead yield comparable results under center lighting, they under-perform in novel lighting scenarios due to the presence of baked center highlights. 
Consistent with expectations, our method facilitates the smoothest time-varying \ac{svBRDF} evolution.

\mysubsection{User Study}{user_study}

\newcommand{\subjects}{Ss\xspace}

In the absence of perceptual metrics for temporal textures, we also quantitatively evaluate our method through a user study. 
Subjects (\subjects) were presented with a reference dynamic texture, along with the dynamic texture results produced by six different methods, arranged in a random layout using a six-alternative choice format. 
In separate trials, participants were asked to identify which alternative was the most: 1) realistic, 2) coherent, and 3) diverse (the diversity question was only for \ac{svBRDF}, as diversity is inherent in all RGB baselines).
For \ac{svBRDF}, results shown to \subjects were relit by the novel flash lighting at the bottom-right corner.
$N=15$ anonymous \subjects were recruited using social media from a university context, but likely na\"ive in respect to the purpose of the study and completed a web form displaying one exemplar with the questions at a time.

In RGB overall realism, 66\,\% preferred our method (\refTab{user_study}), followed by 21\,\% for Two-stream (PF).
The advantage is similar in coherence with 78\,\% vs 7\,\% for the next-best method.

For BRDF overall realism is still higher for state-of-the-art static methods such as MatFusion at 40\,\% as well as Lookahead at 38\,\%, but followed by ours at 25\,\%.
All these BRDF baselines (except \citet{henzler2021generative}), however, are non-generative, and aim to reconstruct the original input, while we produce a a generative texture model, 
It is confirmed in the study, where our diversity is preferred by 45\,\% of the \subjects and only by 18\,\% for the next-best method, MatFusion. 
For coherence, MatFusion and Lookahead are also better at 36\,\% and 31\,\%, respectively, followed by ours at 28\,\%. 
However, the coherence of non-generative models is still just a replication of the original video and relies on the number of frames given. 

\newcolumntype{P}{r<{\,\%}}

\begin{table}[]
    \centering
    \caption{User study aggregates. Preference agreement in percentage.}
    \label{tab:user_study}
    \begin{tabular}{rrPPP}
        \toprule
        &
        &
        \multicolumn1c{Realism}&
        \multicolumn1c{Coher.}&
        \multicolumn1c{Divers.}\\
        \midrule
\multirow{6}{*}{\rotatebox[origin=c]{90}{RGB}}&
Gatys & 4.55 & 1.82 \\
&Ulyanov & 0.61 & 0.00 \\
&Two-stream & 21.52 & 8.79 \\
&DyNCA & 3.33 & 5.76 \\
&SinFusion & 3.64 & 4.85 \\
&Ours & \winner{66.36} & \winner{78.79} \\
\midrule
\multirow{6}{*}{\rotatebox[origin=c]{90}{svBRDF}}&
MaterialGAN & 0.00 & 0.00 & 3.17 \\
&Henzler & 0.00 & 0.00 & 8.25 \\
&MatFusion & \winner{40.32} & \winner{36.19} & 20.00 \\
&Lookahead & 30.79 & 31.43 & 18.10 \\
&Deschaintre & 3.81 & 3.81 & 5.40 \\
&Ours & 25.08 & 28.57 & \winner{45.08} \\
        \bottomrule
        \end{tabular}
\end{table}

\mysection{Discussion}{Disscussion}

\mysubsection{Temporal Coherence}{temporal_coherence_discussion}
We highlight in \refFig{MainComparisons_2} that our method inherently ensures consistency between frames for time-varying appearance via the \ac{ODE} flow, a feature that frame-wise methods cannot achieve.
As observed in the results of frame-wise methods, individual bubbles are randomly jumping between positions across consecutive frames, leading to noticeable flickering--despite the correct growth in bubble size.
While matching the motion somewhat alleviates this issue, it does not resolve the jumps.
In contrast, our method provides a smooth, continuously enlarging trajectory

\myfigure{MainComparisons_2}{The visualization of flickering in frame-wise methods.
We use data \material{Enlarging Bubbles}, bubbles of increasing size.
We show four progressive time points that \emph{average} all previous frames akin to motion blur.
We see that our method produces enlarging bubbles that are consistent over time, while others produce and erase many bubbles, leading to flickers, seen from the overlap of many (\ie ``short-lived'') bubbles.
}
\mysubsection{Architecture Details}{network_details}

\mymath{\scaling}{\sigma}
\mymath{\bias}{\mu}
\mymath{\timeEmbedding}{T}
\paragraph{Time conditioning}
The \ac{AdaGN} is a standard \ac{GN} with time-dependent de-normalization, incorporating the temporal information into the UNet:
$
    \mathtt{AdaGN}(x, \timeEmbedding)=\scaling(\timeEmbedding)\mathtt{GN}(x) + \bias(\timeEmbedding),
$
where the scaling \scaling and the bias \bias are learnable linear projections of the temporal embedding $\timeEmbedding$, which is given as the sinusoidal encoding \timeEncoder \cite{mildenhall2020nerf} of the time input \timeCoord with another \texttt{MLP}:
$
    \timeEmbedding = \mathtt{MLP}(\timeEncoder(\timeCoord))
    .
$
We also experiment with other conditioning methods, \eg directly concatenating the time embedding as extra channels to the input, but have similar observations as \cite{dhariwal2021diffusion,peebles2023scalable}, that \ac{AdaGN} also works better.

\paragraph{Attention}
\myfigure{attention_comp}{Using \ac{ODE} to model \material{Color-changing Leaves} data, with and without attention. We showcase one middle frame highlighting the differences. In the exemplar, redness expands from a specific location, while the generated leaf texture turns red spatial-uniformly if without the attention module.}

We found attention layers to work well in RGB, as ablated in \refFig{attention_comp}, but to be less useful for \ac{svBRDF} where these layers were omitted.

\paragraph{Boundedness}
Our experiments reveal that for challenging inputs with abrupt appearance changes, the \ac{ODE} can become stiff during training. 
Due to stiffness, it requires an impractical number of steps, \ie an unreasonably small step size, to solve the \ac{ODE} within the necessary accuracy.
Regularization \cite{kim2021stiff, finlay2020how} can be leveraged to mitigate stiffness, albeit at the cost of performance capability, so we opt not to use this.

Instead, we empirically find that replacing the commonly used Swish activation, which is also by default in our network, with a Sigmoid activation in the second-to-last layer can significantly alleviate this issue.
In contrast to Swish's unbounded range, the Sigmoid explicitly rectifies the magnitude of inputs of the final layer within $[0, 1]$, which indicates the magnitude of the \ac{ODE} vector field is bounded everywhere by the weights of the final layer, independent of the state and time input.
The boundedness works similarly as in \citet{finlay2020how} by enforcing vector field regularity but requires no extra computational cost and achieves better synthesis quality.

\mysubsection{Effect of ODE Steps}{steps_discussion}
\refFig{Steps} analyzes how many steps per unit time are demanded to solve the \acp{ODE} at every time point \timeCoord of the integral domain $[\timeWarmUp, \timeEnd]$.

\myfigure{Steps}{The number of steps (per unit time) required by the adaptive \ac{ODE} solver is plotted at all time points of $[\timeWarmUp, \timeEnd]$. We plot the average number as the curve, and the variance as the shade, over all \acp{ODE} trained on individual data, and for two tasks (texture, and \ac{svBRDF}) respectively.}

The number of necessary solving steps indicates the difficulty of modeling this piece of dynamics with the \ac{ODE} --- the more complex dynamics require modeling by a stiffer \ac{ODE}, necessitating additional steps to solve to a specified level of accuracy.
It is apparent that during the warm-up phase, the \ac{ODE} solver takes more steps to progress through the same amount of time than in the generation phase, \ie denoising is significantly more challenging than evolution for both tasks. 
This justifies that the initial frame should impose more supervision.
Additionally, for this reason, the use of an adaptive \ac{ODE} solver is important.

\mysubsection{Losses}{loss_discussion}
\myfigure{losses}{Training the \ac{ODE} for \material{Painted-metal Rusting} data with different losses. The top row is our method, which first trains via $\distanceInit$ for initialization and then trains via $\distanceLocal$ for realistic synthesis.}

The efficacy of our initialization strategy is further validated in \refFig{losses}, where a comparison between \acp{ODE} trained with and without this initialization confirms its ability to eliminate baked lighting, particularly evident in the case of low roughness.

Our intuition is that these three forms of losses (\distanceGlobal, \distanceLocal, \distanceInit) lie in the spectrum of varying degrees of stationarity introduced.
The original loss, \distanceGlobal, which incorporates no extra stationarity, is employed to directly compare RGB textures, a task that does not involve solving an inverse problem. 
At the other extreme, the initialization loss, \distanceInit, demands the strongest stationarity and is thus used to establish a favorable starting point. 
In between these, the loss \distanceLocal is primarily used for training \ac{svBRDF} \acp{ODE}, as it strikes a balance between enforcing sufficient stationarity and enabling plausible synthesis.

\mysubsection{Difference to Diffusion Models}{diffusion_discussion}
A similarity among our method, DDPM \cite{ho2020denoising}, SMLD \cite{song2019generative}, and Flow-Matching \cite{lipman2023flow} is the use of a neural network to represent the derivative of a function that is integrated.
However, we do not use a score function or denoising likelihood bound as supervision.
We also do not run a diffusion where the final result is the sample (an image or video), but we run a diffusion, where some of the intermediate states, after warm-up, are intermediate frames of a dynamic solution \cite{chen2019neural}.
Yet, there are commonalities to concepts from diffusion models: using a latent space as in latent diffusion \cite{rombach2022high}.
But then again this is different, as latent diffusion uses a pre-made latent image space, while we have no such step and the latent space to perform dynamics in is learned jointly.

\mysubsection{Adjoint Training}{adjoint_discussion}
Chen \etal propose a memory-efficient method \shortcite{chen2019neural} to acquire gradients through the \ac{ODE} solver, which solves an adjoint \ac{ODE} backward.  
However, the memory is not a bottleneck of training since we have a relatively small model size of around $500$K parameters.
We choose to differentiate through the internal operations of the \ac{ODE} solver to compute exact gradients via back-propagation.

\mysubsection{Performance}{performance_discussion}

Our trained \acp{ODE} can synthesize diverse samples in arbitrary spatial sizes at real-time speeds during inference

On average, our main experiments require 149.7 and 182.2 steps to integrate the entire time range (from \timeWarmUp to \timeEnd) for our RGB \acp{ODE} at $128^2$ resolution and \ac{BRDF} \acp{ODE} at $256^2$ resolution, respectively. 
This translates to merely 1.5 and 1.8 \acp{NFE} per frame.

We test the average wall-clock integration time on an Nvidia 4090 GPU, which reports 0.100, 0.245, and 1.250 seconds for dynamic RGB synthesis at $128^2$, $256^2$, and $512^2$ resolutions, and 0.052, 0.146, and 0.746 seconds for dynamic BRDF synthesis at these resolutions.
If we sample 100 frames in the trajectory, it equates to at least 80FPS.

We summarize how compute time and space requirements relate to our baselines in \refTab{Performance}. 
Our method is efficient in both.

\begin{table}
\centering
\caption{Performance characteristics of different methods.
$^1$ The space unit is the number of network parameters.
$^2$ The time unit is the number of \acp{NFE} to produce a single frame of texture video outputs.
Methods can overfit to one exemplar, train on a large dataset to build a prior, or do both, as labeled in Analysis.
In inference, methods can perform a full optimization or execute the forward pass of a specific neural network, to generate each frame.
}
\label{tab:Performance}
\begin{tabular}{lllrr}
\toprule
& Analysis & Inference & Space$^1$ & Time$^2$ \\
\midrule
Gatys & Overfit & Optimi. & N/A & N/A \\
Ulyanov & Overfit & CNN & 0.56\,M & 1 \\
Two-stream & Overfit & Optimi. & N/A & N/A \\
DyNCA & Overfit & NCA & 0.01\,M & 64 \\
SinFusion & Overfit & Diffusion & 10\,M & 53 \\
Ours & Overfit & ODE & 0.56\,M & 1.5 \\
\midrule
MaterialGAN & Train + overfit & CNN + Opt. & 30\,M & N/A \\
Henzler & Train + overfit & CNN & 28\,M & 1 \\
MatFusion & Train & Diffusion & 256\,M & 20 \\
Lookahead & Train + overfit & CNN & 81\,M & 1 \\
Deschaintre & Train & CNN & 80\,M & 1 \\
Ours & Overfit & ODE & 0.50\,M & 1.8 \\
\bottomrule
\end{tabular}
\end{table}

\mysubsection{Failure Cases}{failure_cases}
For some trials, our method fails to get the most agreements in our user study. Those failure cases more or less violate our assumption, such as exemplar \material{Moving Cloud} which also features in obvious motion flow. 
However, it further consolidates the focus of our work: time-varying visual statistics is one insufficiently explored type of dynamic textures and deserves dedicated attention. 
The previous assumption of motion and our assumption of time-varying visual statistics are not mutually conflicting; instead, they complement each other.
To model a wider spectrum of dynamic textures, our method is orthogonal and can be extended to a motion stream, and vice versa.

\paragraph{Non-stationarity}
We operate under an assumption of spatial stationarity, particularly for the \ac{svBRDF} task, where we rely on this assumption to address the under-constrained issue.
However, the real-world data usually does not strictly adhere to this assumption.
For instance, the exemplar \material{Salt Crystallizing} demonstrates the crystal forming in a specific spatial order, which our method fails to reproduce.
Instead, it places crystal fragments uniformly across all regions, resulting in an artificial appearance.

Resolving this assumption, not being tackled in this work, could be an immediate next step, especially given the extensive research into non-stationary static appearance modeling. 
Diffusion models have prevailed in natural image synthesis, and \ac{svBRDF} acquisition adopts the supervision of synthetic maps and even the generative prior of natural images to deal with the under-constrained issue.
We believe that modeling time-varying appearances in a latent space of non-stationary images using \acp{ODE} represents a promising avenue for future research.

\mycfigure{Comparisons_SVBRDF}{
Comparative results of our method and three baselines for two synthetic exemplars (a specular \material{Painted-metal Rusting} and a diffuse \material{Leather Aging}). 
We concatenate three timesteps from left to right for each dynamic video.
The first row displays the exemplar for each data, starting on the left with the exemplar input with center flash lighting, followed by its ground truth relightings used for evaluation, under two novel flashlights at the top and the bottom-right. 
Results from four methods are shown in the next four rows.
For each method, we present the result of \ac{svBRDF} maps in the first column and relight them with the corresponding novel light conditions in the next two columns.
The layout of the \ac{svBRDF} maps follows the format described in \refFig{DynamicSVBRDF}.
}

\mysection{Conclusion}{Conclusion}
We propose neural differential appearance equations, the main idea being that we learn an ODE that does not generate a final sample, but an initial sample and then drives this sample forward in time, so that the intermediate steps comply with the evolving dynamics and appearance of the exemplar.
We apply this idea to two forms of appearance, RGB textures and \ac{svBRDF} textures.
Previous methods have relied on motion and/or assumed static appearance statistics and failed to synthesize coherent temporal variations of appearance.
Our method offers to reconstruct \acp{svBRDF} of a certain model, enabling relighting.
We inherit the limitations of the shading model used, and future work shall explore effects neglected such as scattering or parallax.
Our results are of competitive visual quality and can be produced efficiently, for applications such as computer games, for online generation of training data, or in the inner loop of yet another optimizer \eg in texture segmentation.

In future work, we would like to apply this idea of warm-up-generation-and-evolve of time-varying content using neural \acp{ODE} to other content such as stereo or light fields, to 3D meshes or 3D point clouds, to latent codes or entire neural representations.

\begin{acks}
This project was supported by Meta Reality Labs, Grant Nr. 583589.
We would also like to thank the constructive comments from the anonymous reviewers and Michael Fischer. 
\end{acks}

\bibliographystyle{ACM-Reference-Format}
\bibliography{paper}


\begin{thebibliography}{78}


\ifx \showCODEN    \undefined \def \showCODEN     #1{\unskip}     \fi
\ifx \showDOI      \undefined \def \showDOI       #1{#1}\fi
\ifx \showISBNx    \undefined \def \showISBNx     #1{\unskip}     \fi
\ifx \showISBNxiii \undefined \def \showISBNxiii  #1{\unskip}     \fi
\ifx \showISSN     \undefined \def \showISSN      #1{\unskip}     \fi
\ifx \showLCCN     \undefined \def \showLCCN      #1{\unskip}     \fi
\ifx \shownote     \undefined \def \shownote      #1{#1}          \fi
\ifx \showarticletitle \undefined \def \showarticletitle #1{#1}   \fi
\ifx \showURL      \undefined \def \showURL       {\relax}        \fi
\providecommand\bibfield[2]{#2}
\providecommand\bibinfo[2]{#2}
\providecommand\natexlab[1]{#1}
\providecommand\showeprint[2][]{arXiv:#2}

\bibitem[Aittala et~al\mbox{.}(2016)]%
        {aittala2016reflectance}
\bibfield{author}{\bibinfo{person}{Miika Aittala}, \bibinfo{person}{Timo Aila}, {and} \bibinfo{person}{Jaakko Lehtinen}.} \bibinfo{year}{2016}\natexlab{}.
\newblock \showarticletitle{Reflectance modeling by neural texture synthesis}.
\newblock \bibinfo{journal}{\emph{ACM Trans. Graph.}} \bibinfo{volume}{35}, \bibinfo{number}{4} (\bibinfo{year}{2016}), \bibinfo{pages}{1--13}.
\newblock


\bibitem[Aittala et~al\mbox{.}(2015)]%
        {aittala2015two}
\bibfield{author}{\bibinfo{person}{Miika Aittala}, \bibinfo{person}{Tim Weyrich}, \bibinfo{person}{Jaakko Lehtinen}, {et~al\mbox{.}}} \bibinfo{year}{2015}\natexlab{}.
\newblock \showarticletitle{Two-shot SVBRDF capture for stationary materials.}
\newblock \bibinfo{journal}{\emph{ACM Trans. Graph.}} \bibinfo{volume}{34}, \bibinfo{number}{4} (\bibinfo{year}{2015}), \bibinfo{pages}{110--1}.
\newblock


\bibitem[Arora and Lee(2021)]%
        {arora2021singan}
\bibfield{author}{\bibinfo{person}{Rajat Arora} {and} \bibinfo{person}{Yong~Jae Lee}.} \bibinfo{year}{2021}\natexlab{}.
\newblock \showarticletitle{Singan-gif: Learning a generative video model from a single gif}. In \bibinfo{booktitle}{\emph{Proc. WAC}}. \bibinfo{pages}{1310--1319}.
\newblock


\bibitem[{Blender Foundation}(2024)]%
        {blender2024}
\bibfield{author}{\bibinfo{person}{{Blender Foundation}}.} \bibinfo{year}{2024}\natexlab{}.
\newblock \bibinfo{title}{Blender}.
\newblock \bibinfo{howpublished}{\url{https://www.blender.org}}.
\newblock
\newblock
\shownote{Version 4.1}.


\bibitem[Box(1976)]%
        {box1976science}
\bibfield{author}{\bibinfo{person}{George~EP Box}.} \bibinfo{year}{1976}\natexlab{}.
\newblock \showarticletitle{Science and statistics}.
\newblock \bibinfo{journal}{\emph{J Am. Stat. Assoc.}} \bibinfo{volume}{71}, \bibinfo{number}{356} (\bibinfo{year}{1976}), \bibinfo{pages}{791--799}.
\newblock


\bibitem[Brooks et~al\mbox{.}(2022)]%
        {brooks2022generating}
\bibfield{author}{\bibinfo{person}{Tim Brooks}, \bibinfo{person}{Janne Hellsten}, \bibinfo{person}{Miika Aittala}, \bibinfo{person}{Ting-Chun Wang}, \bibinfo{person}{Timo Aila}, \bibinfo{person}{Jaakko Lehtinen}, \bibinfo{person}{Ming-Yu Liu}, \bibinfo{person}{Alexei Efros}, {and} \bibinfo{person}{Tero Karras}.} \bibinfo{year}{2022}\natexlab{}.
\newblock \showarticletitle{Generating Long Videos of Dynamic Scenes}.
\newblock \bibinfo{journal}{\emph{NeurIPS}}  \bibinfo{volume}{35} (\bibinfo{year}{2022}), \bibinfo{pages}{31769--31781}.
\newblock


\bibitem[Callister~Jr(2007)]%
        {callisterjr2007materials}
\bibfield{author}{\bibinfo{person}{William~D Callister~Jr}.} \bibinfo{year}{2007}\natexlab{}.
\newblock \bibinfo{booktitle}{\emph{Materials science and engineering an introduction}}.
\newblock


\bibitem[Chen et~al\mbox{.}(2018)]%
        {chen2019neural}
\bibfield{author}{\bibinfo{person}{Ricky~TQ Chen}, \bibinfo{person}{Yulia Rubanova}, \bibinfo{person}{Jesse Bettencourt}, {and} \bibinfo{person}{David~K Duvenaud}.} \bibinfo{year}{2018}\natexlab{}.
\newblock \showarticletitle{Neural ordinary differential equations}.
\newblock \bibinfo{journal}{\emph{Advances in neural information processing systems}}  \bibinfo{volume}{31} (\bibinfo{year}{2018}).
\newblock


\bibitem[Chen et~al\mbox{.}(2005)]%
        {chen2005visual}
\bibfield{author}{\bibinfo{person}{Yanyun Chen}, \bibinfo{person}{Lin Xia}, \bibinfo{person}{Tien-Tsin Wong}, \bibinfo{person}{Xin Tong}, \bibinfo{person}{Hujun Bao}, \bibinfo{person}{Baining Guo}, {and} \bibinfo{person}{Heung-Yeung Shum}.} \bibinfo{year}{2005}\natexlab{}.
\newblock \showarticletitle{Visual simulation of weathering by $\gamma$-ton tracing}.
\newblock In \bibinfo{booktitle}{\emph{ACM SIGGRAPH}}. \bibinfo{pages}{1127--1133}.
\newblock


\bibitem[Cook and Torrance(1982)]%
        {cook1982reflectance}
\bibfield{author}{\bibinfo{person}{Robert~L Cook} {and} \bibinfo{person}{Kenneth~E. Torrance}.} \bibinfo{year}{1982}\natexlab{}.
\newblock \showarticletitle{A reflectance model for computer graphics}.
\newblock \bibinfo{journal}{\emph{ACM Trans. Graph. (ToG)}} \bibinfo{volume}{1}, \bibinfo{number}{1} (\bibinfo{year}{1982}), \bibinfo{pages}{7--24}.
\newblock


\bibitem[Costantini et~al\mbox{.}(2008)]%
        {costantini2008higher}
\bibfield{author}{\bibinfo{person}{Roberto Costantini}, \bibinfo{person}{Luciano Sbaiz}, {and} \bibinfo{person}{Sabine Susstrunk}.} \bibinfo{year}{2008}\natexlab{}.
\newblock \showarticletitle{Higher Order SVD Analysis for Dynamic Texture Synthesis}.
\newblock \bibinfo{journal}{\emph{IEEE Trans. Image Processing}} \bibinfo{volume}{17}, \bibinfo{number}{1} (\bibinfo{year}{2008}), \bibinfo{pages}{42--52}.
\newblock


\bibitem[Desbenoit et~al\mbox{.}(2004)]%
        {desbenoit2004simulating}
\bibfield{author}{\bibinfo{person}{Brett Desbenoit}, \bibinfo{person}{Eric Galin}, {and} \bibinfo{person}{Samir Akkouche}.} \bibinfo{year}{2004}\natexlab{}.
\newblock \showarticletitle{Simulating and modeling lichen growth}.
\newblock  \bibinfo{volume}{23}, \bibinfo{number}{3} (\bibinfo{year}{2004}), \bibinfo{pages}{341--350}.
\newblock


\bibitem[Deschaintre et~al\mbox{.}(2018)]%
        {deschaintre2018single}
\bibfield{author}{\bibinfo{person}{Valentin Deschaintre}, \bibinfo{person}{Miika Aittala}, \bibinfo{person}{Fredo Durand}, \bibinfo{person}{George Drettakis}, {and} \bibinfo{person}{Adrien Bousseau}.} \bibinfo{year}{2018}\natexlab{}.
\newblock \showarticletitle{Single-image svbrdf capture with a rendering-aware deep network}.
\newblock \bibinfo{journal}{\emph{ACM Trans. Graph. (Proc. SIGGRAPH)}} \bibinfo{volume}{37}, \bibinfo{number}{4} (\bibinfo{year}{2018}), \bibinfo{pages}{1--15}.
\newblock


\bibitem[Dhariwal and Nichol(2021)]%
        {dhariwal2021diffusion}
\bibfield{author}{\bibinfo{person}{Prafulla Dhariwal} {and} \bibinfo{person}{Alexander Nichol}.} \bibinfo{year}{2021}\natexlab{}.
\newblock \showarticletitle{Diffusion models beat gans on image synthesis}.
\newblock \bibinfo{journal}{\emph{NeurIPS}}  \bibinfo{volume}{34} (\bibinfo{year}{2021}), \bibinfo{pages}{8780--8794}.
\newblock


\bibitem[Doretto et~al\mbox{.}(2003)]%
        {doretto2003dynamic}
\bibfield{author}{\bibinfo{person}{Gianfranco Doretto}, \bibinfo{person}{Alessandro Chiuso}, \bibinfo{person}{Ying~Nian Wu}, {and} \bibinfo{person}{Stefano Soatto}.} \bibinfo{year}{2003}\natexlab{}.
\newblock \showarticletitle{Dynamic Textures}.
\newblock \bibinfo{journal}{\emph{Int. J Comp. Vis.}} \bibinfo{volume}{51}, \bibinfo{number}{2} (\bibinfo{year}{2003}), \bibinfo{pages}{91--109}.
\newblock


\bibitem[Dorsey et~al\mbox{.}(2005)]%
        {dorsey2005modeling}
\bibfield{author}{\bibinfo{person}{Julie Dorsey}, \bibinfo{person}{Alan Edelman}, \bibinfo{person}{Henrik~Wann Jensen}, \bibinfo{person}{Justin Legakis}, {and} \bibinfo{person}{Hans~Köhling Pedersen}.} \bibinfo{year}{2005}\natexlab{}.
\newblock \showarticletitle{Modeling and rendering of weathered stone}.
\newblock In \bibinfo{booktitle}{\emph{ACM SIGGRAPH 2005 Courses}}. \bibinfo{pages}{4--es}.
\newblock


\bibitem[Dorsey and Hanrahan(2006)]%
        {dorsey2006modeling}
\bibfield{author}{\bibinfo{person}{Julie Dorsey} {and} \bibinfo{person}{Pat Hanrahan}.} \bibinfo{year}{2006}\natexlab{}.
\newblock \showarticletitle{Modeling and rendering of metallic patinas}.
\newblock In \bibinfo{booktitle}{\emph{ACM SIGGRAPH Courses}}. \bibinfo{pages}{2--es}.
\newblock


\bibitem[Dorsey et~al\mbox{.}(1996)]%
        {dorsey1996flow}
\bibfield{author}{\bibinfo{person}{Julie Dorsey}, \bibinfo{person}{Hans~Köhling Pedersen}, {and} \bibinfo{person}{Pat Hanrahan}.} \bibinfo{year}{1996}\natexlab{}.
\newblock \showarticletitle{Flow and changes in appearance}. In \bibinfo{booktitle}{\emph{SIGGRAPH}}. \bibinfo{pages}{411--420}.
\newblock


\bibitem[Dupont et~al\mbox{.}(2019)]%
        {dupont2019augmented}
\bibfield{author}{\bibinfo{person}{Emilien Dupont}, \bibinfo{person}{Arnaud Doucet}, {and} \bibinfo{person}{Yee~Whye Teh}.} \bibinfo{year}{2019}\natexlab{}.
\newblock \showarticletitle{Augmented Neural ODEs}. In \bibinfo{booktitle}{\emph{NeurIPS}}, Vol.~\bibinfo{volume}{32}.
\newblock


\bibitem[Finlay et~al\mbox{.}(2020)]%
        {finlay2020how}
\bibfield{author}{\bibinfo{person}{Chris Finlay}, \bibinfo{person}{Joern-Henrik Jacobsen}, \bibinfo{person}{Levon Nurbekyan}, {and} \bibinfo{person}{Adam Oberman}.} \bibinfo{year}{2020}\natexlab{}.
\newblock \showarticletitle{How to Train Your Neural ODE: The World of Jacobian and Kinetic Regularization}. In \bibinfo{booktitle}{\emph{ICML}}. \bibinfo{pages}{3154--3164}.
\newblock


\bibitem[Fischer and Ritschel(2022)]%
        {fischer2022metappearance}
\bibfield{author}{\bibinfo{person}{Michael Fischer} {and} \bibinfo{person}{Tobias Ritschel}.} \bibinfo{year}{2022}\natexlab{}.
\newblock \showarticletitle{Metappearance: Meta-learning for visual appearance reproduction}.
\newblock \bibinfo{journal}{\emph{ACM Trans. Graph. (Proc. SIGGRAPH Asia)}} \bibinfo{volume}{41}, \bibinfo{number}{6} (\bibinfo{year}{2022}).
\newblock


\bibitem[Funke et~al\mbox{.}(2017)]%
        {funke2017synthesising}
\bibfield{author}{\bibinfo{person}{Christina~M. Funke}, \bibinfo{person}{Leon~A. Gatys}, \bibinfo{person}{Alexander~S. Ecker}, {and} \bibinfo{person}{Matthias Bethge}.} \bibinfo{year}{2017}\natexlab{}.
\newblock \bibinfo{booktitle}{\emph{Synthesising Dynamic Textures Using Convolutional Neural Networks}}.
\newblock
\showeprint[arxiv]{1702.07006}~[cs]


\bibitem[Gao et~al\mbox{.}(2019)]%
        {gao2019deep}
\bibfield{author}{\bibinfo{person}{Duan Gao}, \bibinfo{person}{Xiao Li}, \bibinfo{person}{Yue Dong}, \bibinfo{person}{Pieter Peers}, \bibinfo{person}{Kun Xu}, {and} \bibinfo{person}{Xin Tong}.} \bibinfo{year}{2019}\natexlab{}.
\newblock \showarticletitle{Deep inverse rendering for high-resolution SVBRDF estimation from an arbitrary number of images.}
\newblock \bibinfo{journal}{\emph{ACM Trans. Graph.}} \bibinfo{volume}{38}, \bibinfo{number}{4} (\bibinfo{year}{2019}), \bibinfo{pages}{134--1}.
\newblock


\bibitem[Gatys et~al\mbox{.}(2015)]%
        {gatys2015texture}
\bibfield{author}{\bibinfo{person}{Leon Gatys}, \bibinfo{person}{Alexander~S Ecker}, {and} \bibinfo{person}{Matthias Bethge}.} \bibinfo{year}{2015}\natexlab{}.
\newblock \showarticletitle{Texture Synthesis Using Convolutional Neural Networks}. In \bibinfo{booktitle}{\emph{NeurIPS}}, Vol.~\bibinfo{volume}{28}.
\newblock


\bibitem[Georgoulis et~al\mbox{.}(2017)]%
        {georgoulis2017reflectance}
\bibfield{author}{\bibinfo{person}{Stamatios Georgoulis}, \bibinfo{person}{Konstantinos Rematas}, \bibinfo{person}{Tobias Ritschel}, \bibinfo{person}{Efstratios Gavves}, \bibinfo{person}{Mario Fritz}, \bibinfo{person}{Luc Van~Gool}, {and} \bibinfo{person}{Tinne Tuytelaars}.} \bibinfo{year}{2017}\natexlab{}.
\newblock \showarticletitle{Reflectance and natural illumination from single-material specular objects using deep learning}.
\newblock \bibinfo{journal}{\emph{IEEE Trans. PAMI}} \bibinfo{volume}{40}, \bibinfo{number}{8} (\bibinfo{year}{2017}), \bibinfo{pages}{1932--1947}.
\newblock


\bibitem[Gobron and Chiba(2001)]%
        {gobron2001crack}
\bibfield{author}{\bibinfo{person}{St\'ephane Gobron} {and} \bibinfo{person}{Norishige Chiba}.} \bibinfo{year}{2001}\natexlab{}.
\newblock \showarticletitle{Crack pattern simulation based on {3D} surface cellular automata}.
\newblock \bibinfo{journal}{\emph{The Visual Computer}}  \bibinfo{volume}{17} (\bibinfo{year}{2001}), \bibinfo{pages}{287--309}.
\newblock


\bibitem[Gu et~al\mbox{.}(2006)]%
        {gu2006time}
\bibfield{author}{\bibinfo{person}{Jinwei Gu}, \bibinfo{person}{Chien-I Tu}, \bibinfo{person}{Ravi Ramamoorthi}, \bibinfo{person}{Peter Belhumeur}, \bibinfo{person}{Wojciech Matusik}, {and} \bibinfo{person}{Shree Nayar}.} \bibinfo{year}{2006}\natexlab{}.
\newblock \showarticletitle{Time-varying surface appearance: acquisition, modeling and rendering}.
\newblock \bibinfo{journal}{\emph{ACM Trans. Graph.}} \bibinfo{volume}{25}, \bibinfo{number}{3} (\bibinfo{year}{2006}), \bibinfo{pages}{762--771}.
\newblock


\bibitem[Guo et~al\mbox{.}(2020)]%
        {guo2020materialgan}
\bibfield{author}{\bibinfo{person}{Yu Guo}, \bibinfo{person}{Cameron Smith}, \bibinfo{person}{Milo\v{s} Ha\v{s}an}, \bibinfo{person}{Kalyan Sunkavalli}, {and} \bibinfo{person}{Shuang Zhao}.} \bibinfo{year}{2020}\natexlab{}.
\newblock \showarticletitle{MaterialGAN: reflectance capture using a generative SVBRDF model}.
\newblock \bibinfo{journal}{\emph{ACM Trans. Graph.}} \bibinfo{volume}{39}, \bibinfo{number}{6} (\bibinfo{year}{2020}).
\newblock


\bibitem[Hadji and Wildes(2018)]%
        {hadji2018new}
\bibfield{author}{\bibinfo{person}{Isma Hadji} {and} \bibinfo{person}{Richard~P. Wildes}.} \bibinfo{year}{2018}\natexlab{}.
\newblock \showarticletitle{A New Large Scale Dynamic Texture Dataset with Application to {ConvNet} Understanding}. In \bibinfo{booktitle}{\emph{ECCV}}. \bibinfo{pages}{334--351}.
\newblock


\bibitem[Harrington et~al\mbox{.}(2022)]%
        {harrington2022exploring}
\bibfield{author}{\bibinfo{person}{Anne Harrington}, \bibinfo{person}{Vasha DuTell}, \bibinfo{person}{Ayush Tewari}, \bibinfo{person}{Mark Hamilton}, \bibinfo{person}{Simon Stent}, \bibinfo{person}{Ruth Rosenholtz}, {and} \bibinfo{person}{William~T Freeman}.} \bibinfo{year}{2022}\natexlab{}.
\newblock \showarticletitle{Exploring perceptual straightness in learned visual representations}. In \bibinfo{booktitle}{\emph{ICLR}}.
\newblock


\bibitem[Heitz(2014)]%
        {heitz2014understanding}
\bibfield{author}{\bibinfo{person}{Eric Heitz}.} \bibinfo{year}{2014}\natexlab{}.
\newblock \showarticletitle{Understanding the masking-shadowing function in microfacet-based BRDFs}.
\newblock \bibinfo{journal}{\emph{J Comp. Graph. Techniques}} \bibinfo{volume}{3}, \bibinfo{number}{2} (\bibinfo{year}{2014}), \bibinfo{pages}{32--91}.
\newblock


\bibitem[Heitz et~al\mbox{.}(2023)]%
        {heitz2023iterative}
\bibfield{author}{\bibinfo{person}{Eric Heitz}, \bibinfo{person}{Laurent Belcour}, {and} \bibinfo{person}{Thomas Chambon}.} \bibinfo{year}{2023}\natexlab{}.
\newblock \showarticletitle{Iterative $\alpha$-(de) blending: A minimalist deterministic diffusion model}. In \bibinfo{booktitle}{\emph{ACM SIGGRAPH 2023 Conference Proceedings}}. \bibinfo{pages}{1--8}.
\newblock


\bibitem[Heitz et~al\mbox{.}(2021)]%
        {heitz2021sliced}
\bibfield{author}{\bibinfo{person}{Eric Heitz}, \bibinfo{person}{Kenneth Vanhoey}, \bibinfo{person}{Thomas Chambon}, {and} \bibinfo{person}{Laurent Belcour}.} \bibinfo{year}{2021}\natexlab{}.
\newblock \showarticletitle{A Sliced Wasserstein Loss for Neural Texture Synthesis}. \bibinfo{pages}{9412--9420}.
\newblock


\bibitem[H{\'e}naff et~al\mbox{.}(2019)]%
        {henaff2019perceptual}
\bibfield{author}{\bibinfo{person}{Olivier~J H{\'e}naff}, \bibinfo{person}{Robbe~LT Goris}, {and} \bibinfo{person}{Eero~P Simoncelli}.} \bibinfo{year}{2019}\natexlab{}.
\newblock \showarticletitle{Perceptual straightening of natural videos}.
\newblock \bibinfo{journal}{\emph{Nature Neuroscience}} \bibinfo{volume}{22}, \bibinfo{number}{6} (\bibinfo{year}{2019}), \bibinfo{pages}{984--991}.
\newblock


\bibitem[Henzler et~al\mbox{.}(2021)]%
        {henzler2021generative}
\bibfield{author}{\bibinfo{person}{Philipp Henzler}, \bibinfo{person}{Valentin Deschaintre}, \bibinfo{person}{Niloy~J Mitra}, {and} \bibinfo{person}{Tobias Ritschel}.} \bibinfo{year}{2021}\natexlab{}.
\newblock \showarticletitle{Generative Modelling of BRDF Textures from Flash Images}.
\newblock \bibinfo{journal}{\emph{ACM Trans. Graph. (Proc. SIGGRAPH Asia)}} \bibinfo{volume}{40}, \bibinfo{number}{6} (\bibinfo{year}{2021}).
\newblock


\bibitem[Ho et~al\mbox{.}(2020)]%
        {ho2020denoising}
\bibfield{author}{\bibinfo{person}{Jonathan Ho}, \bibinfo{person}{Ajay Jain}, {and} \bibinfo{person}{Pieter Abbeel}.} \bibinfo{year}{2020}\natexlab{}.
\newblock \showarticletitle{Denoising Diffusion Probabilistic Models}. In \bibinfo{booktitle}{\emph{NeurIPS}}, Vol.~\bibinfo{volume}{33}. \bibinfo{pages}{6840--6851}.
\newblock


\bibitem[Hsu and Wong(1995)]%
        {hsu1995simulating}
\bibfield{author}{\bibinfo{person}{Siu-Chi Hsu} {and} \bibinfo{person}{Tien-Tsin Wong}.} \bibinfo{year}{1995}\natexlab{}.
\newblock \showarticletitle{Simulating dust accumulation}.
\newblock \bibinfo{journal}{\emph{IEEE Computer Graphics and Applications}} \bibinfo{volume}{15}, \bibinfo{number}{1} (\bibinfo{year}{1995}), \bibinfo{pages}{18--22}.
\newblock


\bibitem[Kider~Jr et~al\mbox{.}(2011)]%
        {kider2011fruit}
\bibfield{author}{\bibinfo{person}{Joseph~T Kider~Jr}, \bibinfo{person}{Samantha Raja}, {and} \bibinfo{person}{Norman~I Badler}.} \bibinfo{year}{2011}\natexlab{}.
\newblock \showarticletitle{Fruit senescence and decay simulation}. In \bibinfo{booktitle}{\emph{Comp. Graph. Forum}}, Vol.~\bibinfo{volume}{30}. \bibinfo{pages}{257--266}.
\newblock


\bibitem[Kim et~al\mbox{.}(2021)]%
        {kim2021stiff}
\bibfield{author}{\bibinfo{person}{Suyong Kim}, \bibinfo{person}{Weiqi Ji}, \bibinfo{person}{Sili Deng}, \bibinfo{person}{Yingbo Ma}, {and} \bibinfo{person}{Christopher Rackauckas}.} \bibinfo{year}{2021}\natexlab{}.
\newblock \showarticletitle{Stiff Neural Ordinary Differential Equations}.
\newblock \bibinfo{journal}{\emph{Chaos: An Interdisciplinary Journal of Nonlinear Science}} \bibinfo{volume}{31}, \bibinfo{number}{9} (\bibinfo{year}{2021}), \bibinfo{pages}{093122}.
\newblock
\showeprint[arxiv]{2103.15341}~[cs, math]


\bibitem[Li et~al\mbox{.}(2017)]%
        {li2017modeling}
\bibfield{author}{\bibinfo{person}{Xiao Li}, \bibinfo{person}{Yue Dong}, \bibinfo{person}{Pieter Peers}, {and} \bibinfo{person}{Xin Tong}.} \bibinfo{year}{2017}\natexlab{}.
\newblock \showarticletitle{Modeling surface appearance from a single photograph using self-augmented convolutional neural networks}.
\newblock \bibinfo{journal}{\emph{ACM Trans. Graph.}} \bibinfo{volume}{36}, \bibinfo{number}{4} (\bibinfo{year}{2017}), \bibinfo{pages}{1--11}.
\newblock


\bibitem[Lipman et~al\mbox{.}(2023)]%
        {lipman2023flow}
\bibfield{author}{\bibinfo{person}{Yaron Lipman}, \bibinfo{person}{Ricky T.~Q. Chen}, \bibinfo{person}{Heli Ben{-}Hamu}, \bibinfo{person}{Maximilian Nickel}, {and} \bibinfo{person}{Matthew Le}.} \bibinfo{year}{2023}\natexlab{}.
\newblock \showarticletitle{Flow Matching for Generative Modeling}. In \bibinfo{booktitle}{\emph{The Eleventh International Conference on Learning Representations, {ICLR} 2023, Kigali, Rwanda, May 1-5, 2023}}. \bibinfo{publisher}{OpenReview.net}.
\newblock


\bibitem[Lombardi and Nishino(2015)]%
        {lombardi2015reflectance}
\bibfield{author}{\bibinfo{person}{Stephen Lombardi} {and} \bibinfo{person}{Ko Nishino}.} \bibinfo{year}{2015}\natexlab{}.
\newblock \showarticletitle{Reflectance and illumination recovery in the wild}.
\newblock \bibinfo{journal}{\emph{IEEE Trans. PAMI}} \bibinfo{volume}{38}, \bibinfo{number}{1} (\bibinfo{year}{2015}), \bibinfo{pages}{129--141}.
\newblock


\bibitem[Marschner(1998)]%
        {marschner1998inverse}
\bibfield{author}{\bibinfo{person}{Stephen~Robert Marschner}.} \bibinfo{year}{1998}\natexlab{}.
\newblock \bibinfo{booktitle}{\emph{Inverse rendering for computer graphics}}.
\newblock


\bibitem[Matusik(2003)]%
        {matusik2003data}
\bibfield{author}{\bibinfo{person}{Wojciech Matusik}.} \bibinfo{year}{2003}\natexlab{}.
\newblock \emph{\bibinfo{title}{A data-driven reflectance model}}.
\newblock \bibinfo{thesistype}{Ph.\,D. Dissertation}. \bibinfo{school}{Massachusetts Institute of Technology}.
\newblock


\bibitem[Merillou et~al\mbox{.}(2001)]%
        {merillou2001corrosion}
\bibfield{author}{\bibinfo{person}{Stephane Merillou}, \bibinfo{person}{Jean-Michel Dischler}, {and} \bibinfo{person}{Djamchid Ghazanfarpour}.} \bibinfo{year}{2001}\natexlab{}.
\newblock \showarticletitle{Corrosion: Simulating and rendering}. In \bibinfo{booktitle}{\emph{Graphics interface}}, Vol.~\bibinfo{volume}{2001}. \bibinfo{pages}{167--174}.
\newblock


\bibitem[M{\'e}rillou and Ghazanfarpour(2008)]%
        {merillou2008survey}
\bibfield{author}{\bibinfo{person}{St{\'e}phane M{\'e}rillou} {and} \bibinfo{person}{Djamchid Ghazanfarpour}.} \bibinfo{year}{2008}\natexlab{}.
\newblock \showarticletitle{A survey of aging and weathering phenomena in computer graphics}.
\newblock \bibinfo{journal}{\emph{Computers \& Graphics}} \bibinfo{volume}{32}, \bibinfo{number}{2} (\bibinfo{year}{2008}), \bibinfo{pages}{159--174}.
\newblock


\bibitem[Mildenhall et~al\mbox{.}(2020)]%
        {mildenhall2020nerf}
\bibfield{author}{\bibinfo{person}{Ben Mildenhall}, \bibinfo{person}{Pratul~P. Srinivasan}, \bibinfo{person}{Matthew Tancik}, \bibinfo{person}{Jonathan~T. Barron}, \bibinfo{person}{Ravi Ramamoorthi}, {and} \bibinfo{person}{Ren Ng}.} \bibinfo{year}{2020}\natexlab{}.
\newblock \showarticletitle{NeRF: Representing Scenes as Neural Radiance Fields for View Synthesis}. In \bibinfo{booktitle}{\emph{ECCV}}. \bibinfo{pages}{405--421}.
\newblock


\bibitem[Mordvintsev et~al\mbox{.}(2021)]%
        {mordvintsev2021texture}
\bibfield{author}{\bibinfo{person}{Alexander Mordvintsev}, \bibinfo{person}{Eyvind Niklasson}, {and} \bibinfo{person}{Ettore Randazzo}.} \bibinfo{year}{2021}\natexlab{}.
\newblock \bibinfo{booktitle}{\emph{Texture Generation with Neural Cellular Automata}}.
\newblock
\showeprint[arxiv]{2105.07299}~[cs]


\bibitem[Mordvintsev et~al\mbox{.}(2020a)]%
        {mordvintsev2020growing}
\bibfield{author}{\bibinfo{person}{Alexander Mordvintsev}, \bibinfo{person}{Ettore Randazzo}, \bibinfo{person}{Eyvind Niklasson}, {and} \bibinfo{person}{Michael Levin}.} \bibinfo{year}{2020}\natexlab{a}.
\newblock \showarticletitle{Growing Neural Cellular Automata}.
\newblock \bibinfo{journal}{\emph{Distill}} \bibinfo{volume}{5}, \bibinfo{number}{2} (\bibinfo{year}{2020}), \bibinfo{pages}{e23}.
\newblock


\bibitem[Mordvintsev et~al\mbox{.}(2020b)]%
        {mordvintsev2020thread}
\bibfield{author}{\bibinfo{person}{Alexander Mordvintsev}, \bibinfo{person}{Ettore Randazzo}, \bibinfo{person}{Eyvind Niklasson}, \bibinfo{person}{Michael Levin}, {and} \bibinfo{person}{Sam Greydanus}.} \bibinfo{year}{2020}\natexlab{b}.
\newblock \showarticletitle{Thread: Differentiable Self-Organizing Systems}.
\newblock \bibinfo{journal}{\emph{Distill}} (\bibinfo{year}{2020}).
\newblock


\bibitem[Nam et~al\mbox{.}(2019)]%
        {nam2019end}
\bibfield{author}{\bibinfo{person}{Seonghyeon Nam}, \bibinfo{person}{Chongyang Ma}, \bibinfo{person}{Menglei Chai}, \bibinfo{person}{William Brendel}, \bibinfo{person}{Ning Xu}, {and} \bibinfo{person}{Seon~Joo Kim}.} \bibinfo{year}{2019}\natexlab{}.
\newblock \showarticletitle{End-To-End Time-Lapse Video Synthesis From a Single Outdoor Image}. \bibinfo{pages}{1409--1418}.
\newblock


\bibitem[Nikankin et~al\mbox{.}(2023)]%
        {nikankin2023sinfusion}
\bibfield{author}{\bibinfo{person}{Yaniv Nikankin}, \bibinfo{person}{Niv Haim}, {and} \bibinfo{person}{Michal Irani}.} \bibinfo{year}{2023}\natexlab{}.
\newblock \showarticletitle{SinFusion: training diffusion models on a single image or video}. In \bibinfo{booktitle}{\emph{ICML}}. Article \bibinfo{articleno}{1090}, \bibinfo{numpages}{16}~pages.
\newblock


\bibitem[Niklasson et~al\mbox{.}(2021)]%
        {niklasson2021self}
\bibfield{author}{\bibinfo{person}{Eyvind Niklasson}, \bibinfo{person}{Alexander Mordvintsev}, \bibinfo{person}{Ettore Randazzo}, {and} \bibinfo{person}{Michael Levin}.} \bibinfo{year}{2021}\natexlab{}.
\newblock \showarticletitle{Self-Organising Textures}.
\newblock \bibinfo{journal}{\emph{Distill}} \bibinfo{volume}{6}, \bibinfo{number}{2} (\bibinfo{year}{2021}), \bibinfo{pages}{10.23915/distill.00027.003}.
\newblock


\bibitem[Pajouheshgar et~al\mbox{.}(2023a)]%
        {pajouheshgar2023mesh}
\bibfield{author}{\bibinfo{person}{Ehsan Pajouheshgar}, \bibinfo{person}{Yitao Xu}, \bibinfo{person}{Alexander Mordvintsev}, \bibinfo{person}{Eyvind Niklasson}, \bibinfo{person}{Tong Zhang}, {and} \bibinfo{person}{Sabine Süsstrunk}.} \bibinfo{year}{2023}\natexlab{a}.
\newblock \bibinfo{booktitle}{\emph{Mesh Neural Cellular Automata}}.
\newblock
\showeprint[arxiv]{2311.02820}~[cs]


\bibitem[Pajouheshgar et~al\mbox{.}(2023b)]%
        {pajouheshgar2023dynca}
\bibfield{author}{\bibinfo{person}{Ehsan Pajouheshgar}, \bibinfo{person}{Yitao Xu}, \bibinfo{person}{Tong Zhang}, {and} \bibinfo{person}{Sabine S{\"u}sstrunk}.} \bibinfo{year}{2023}\natexlab{b}.
\newblock \showarticletitle{Dynca: Real-time dynamic texture synthesis using neural cellular automata}. In \bibinfo{booktitle}{\emph{Proceedings of the IEEE/CVF Conference on Computer Vision and Pattern Recognition}}. \bibinfo{pages}{20742--20751}.
\newblock


\bibitem[Paquette et~al\mbox{.}(2002)]%
        {paquette2002simulation}
\bibfield{author}{\bibinfo{person}{Eric Paquette}, \bibinfo{person}{Pierre Poulin}, {and} \bibinfo{person}{George Drettakis}.} \bibinfo{year}{2002}\natexlab{}.
\newblock \showarticletitle{The simulation of paint cracking and peeling}. In \bibinfo{booktitle}{\emph{Proc. Graphics Interface}}. \bibinfo{pages}{10}.
\newblock


\bibitem[Peebles and Xie(2023)]%
        {peebles2023scalable}
\bibfield{author}{\bibinfo{person}{William Peebles} {and} \bibinfo{person}{Saining Xie}.} \bibinfo{year}{2023}\natexlab{}.
\newblock \showarticletitle{Scalable diffusion models with transformers}. In \bibinfo{booktitle}{\emph{ICCV}}. \bibinfo{pages}{4195--4205}.
\newblock


\bibitem[Péteri et~al\mbox{.}(2010)]%
        {peteri2010dyntex}
\bibfield{author}{\bibinfo{person}{Renaud Péteri}, \bibinfo{person}{Sándor Fazekas}, {and} \bibinfo{person}{Mark~J. Huiskes}.} \bibinfo{year}{2010}\natexlab{}.
\newblock \showarticletitle{DynTex: A Comprehensive Database of Dynamic Textures}.
\newblock \bibinfo{journal}{\emph{Pattern Recognition Letters}} \bibinfo{volume}{31}, \bibinfo{number}{12} (\bibinfo{year}{2010}), \bibinfo{pages}{1627--1632}.
\newblock


\bibitem[Rombach et~al\mbox{.}(2022)]%
        {rombach2022high}
\bibfield{author}{\bibinfo{person}{Robin Rombach}, \bibinfo{person}{Andreas Blattmann}, \bibinfo{person}{Dominik Lorenz}, \bibinfo{person}{Patrick Esser}, {and} \bibinfo{person}{Björn Ommer}.} \bibinfo{year}{2022}\natexlab{}.
\newblock \showarticletitle{High-Resolution Image Synthesis with Latent Diffusion Models}. In \bibinfo{booktitle}{\emph{CVPR}}. \bibinfo{pages}{10674--10685}.
\newblock


\bibitem[Ronneberger et~al\mbox{.}(2015)]%
        {ronneberger2015u}
\bibfield{author}{\bibinfo{person}{Olaf Ronneberger}, \bibinfo{person}{Philipp Fischer}, {and} \bibinfo{person}{Thomas Brox}.} \bibinfo{year}{2015}\natexlab{}.
\newblock \showarticletitle{U-net: Convolutional networks for biomedical image segmentation}. In \bibinfo{booktitle}{\emph{Proc. MICCAI}}. \bibinfo{pages}{234--241}.
\newblock


\bibitem[Sartor and Peers(2023)]%
        {sartor2023matfusion}
\bibfield{author}{\bibinfo{person}{Sam Sartor} {and} \bibinfo{person}{Pieter Peers}.} \bibinfo{year}{2023}\natexlab{}.
\newblock \showarticletitle{Matfusion: a generative diffusion model for {SVRBD} capture}. In \bibinfo{booktitle}{\emph{SIGGRAPH Asia}}. \bibinfo{pages}{1--10}.
\newblock


\bibitem[Schlick(1994)]%
        {schlick1994inexpensive}
\bibfield{author}{\bibinfo{person}{Christophe Schlick}.} \bibinfo{year}{1994}\natexlab{}.
\newblock \showarticletitle{An inexpensive BRDF model for physically-based rendering}.
\newblock  \bibinfo{volume}{13}, \bibinfo{number}{3} (\bibinfo{year}{1994}), \bibinfo{pages}{233--246}.
\newblock


\bibitem[Smith(1967)]%
        {smith1967geometrical}
\bibfield{author}{\bibinfo{person}{Bruce Smith}.} \bibinfo{year}{1967}\natexlab{}.
\newblock \showarticletitle{Geometrical shadowing of a random rough surface}.
\newblock \bibinfo{journal}{\emph{IEEE Trans. Antennas and Propagation}} \bibinfo{volume}{15}, \bibinfo{number}{5} (\bibinfo{year}{1967}), \bibinfo{pages}{668--671}.
\newblock


\bibitem[Sohl-Dickstein et~al\mbox{.}(2015)]%
        {sohldickstein2015deep}
\bibfield{author}{\bibinfo{person}{Jascha Sohl-Dickstein}, \bibinfo{person}{Eric Weiss}, \bibinfo{person}{Niru Maheswaranathan}, {and} \bibinfo{person}{Surya Ganguli}.} \bibinfo{year}{2015}\natexlab{}.
\newblock \showarticletitle{Deep Unsupervised Learning using Nonequilibrium Thermodynamics}. In \bibinfo{booktitle}{\emph{ICML}}, Vol.~\bibinfo{volume}{37}. \bibinfo{pages}{2256--2265}.
\newblock


\bibitem[Song and Ermon(2019)]%
        {song2019generative}
\bibfield{author}{\bibinfo{person}{Yang Song} {and} \bibinfo{person}{Stefano Ermon}.} \bibinfo{year}{2019}\natexlab{}.
\newblock \showarticletitle{Generative Modeling by Estimating Gradients of the Data Distribution}. In \bibinfo{booktitle}{\emph{NeurIPS}}, Vol.~\bibinfo{volume}{32}.
\newblock


\bibitem[Sun et~al\mbox{.}(2007)]%
        {sun2007time}
\bibfield{author}{\bibinfo{person}{Bo Sun}, \bibinfo{person}{Kalyan Sunkavalli}, \bibinfo{person}{Ravi Ramamoorthi}, \bibinfo{person}{Peter~N Belhumeur}, {and} \bibinfo{person}{Shree~K Nayar}.} \bibinfo{year}{2007}\natexlab{}.
\newblock \showarticletitle{Time-varying {BRDFs}}.
\newblock \bibinfo{journal}{\emph{IEEE TVCG}} \bibinfo{volume}{13}, \bibinfo{number}{03} (\bibinfo{year}{2007}), \bibinfo{pages}{595--609}.
\newblock


\bibitem[Tesfaldet et~al\mbox{.}(2018)]%
        {tesfaldet2018two}
\bibfield{author}{\bibinfo{person}{Matthew Tesfaldet}, \bibinfo{person}{Marcus~A. Brubaker}, {and} \bibinfo{person}{Konstantinos~G. Derpanis}.} \bibinfo{year}{2018}\natexlab{}.
\newblock \showarticletitle{Two-Stream Convolutional Networks for Dynamic Texture Synthesis}.
\newblock  (\bibinfo{year}{2018}), \bibinfo{pages}{6703--6712}.
\newblock


\bibitem[Turing(1952)]%
        {turing1952chemical}
\bibfield{author}{\bibinfo{person}{A.~M. Turing}.} \bibinfo{year}{1952}\natexlab{}.
\newblock \showarticletitle{The chemical basis of morphogenesis}.
\newblock \bibinfo{journal}{\emph{Sciences}} \bibinfo{volume}{237}, \bibinfo{number}{641} (\bibinfo{year}{1952}), \bibinfo{pages}{37--72}.
\newblock


\bibitem[Turk(1991)]%
        {turk1991generating}
\bibfield{author}{\bibinfo{person}{Greg Turk}.} \bibinfo{year}{1991}\natexlab{}.
\newblock \showarticletitle{Generating textures on arbitrary surfaces using reaction-diffusion}.
\newblock \bibinfo{journal}{\emph{ACM Siggraph Computer Graphics}} \bibinfo{volume}{25}, \bibinfo{number}{4} (\bibinfo{year}{1991}), \bibinfo{pages}{289--298}.
\newblock


\bibitem[Ulyanov et~al\mbox{.}(2017)]%
        {ulyanov2017improved}
\bibfield{author}{\bibinfo{person}{Dmitry Ulyanov}, \bibinfo{person}{Andrea Vedaldi}, {and} \bibinfo{person}{Victor Lempitsky}.} \bibinfo{year}{2017}\natexlab{}.
\newblock \showarticletitle{Improved texture networks: Maximizing quality and diversity in feed-forward stylization and texture synthesis}. In \bibinfo{booktitle}{\emph{CVPR}}. \bibinfo{pages}{6924--6932}.
\newblock


\bibitem[Wang et~al\mbox{.}(2006)]%
        {wang2006appearance}
\bibfield{author}{\bibinfo{person}{Jiaping Wang}, \bibinfo{person}{Xin Tong}, \bibinfo{person}{Stephen Lin}, \bibinfo{person}{Minghao Pan}, \bibinfo{person}{Chao Wang}, \bibinfo{person}{Hujun Bao}, \bibinfo{person}{Baining Guo}, {and} \bibinfo{person}{Heung-Yeung Shum}.} \bibinfo{year}{2006}\natexlab{}.
\newblock \showarticletitle{Appearance manifolds for modeling time-variant appearance of materials}.
\newblock \bibinfo{journal}{\emph{ACM Trans. Graph.}} \bibinfo{volume}{25}, \bibinfo{number}{3} (\bibinfo{year}{2006}), \bibinfo{pages}{754--761}.
\newblock


\bibitem[Witkin and Kass(1991)]%
        {witkin1991reaction}
\bibfield{author}{\bibinfo{person}{Andrew Witkin} {and} \bibinfo{person}{Michael Kass}.} \bibinfo{year}{1991}\natexlab{}.
\newblock \showarticletitle{Reaction-diffusion textures}. In \bibinfo{booktitle}{\emph{SIGGRAPH}}. \bibinfo{pages}{299--308}.
\newblock


\bibitem[Xie et~al\mbox{.}(2017)]%
        {xie2017synthesizing}
\bibfield{author}{\bibinfo{person}{Jianwen Xie}, \bibinfo{person}{Song-Chun Zhu}, {and} \bibinfo{person}{Ying~Nian Wu}.} \bibinfo{year}{2017}\natexlab{}.
\newblock \showarticletitle{Synthesizing Dynamic Patterns by Spatial-Temporal Generative ConvNet}. In \bibinfo{booktitle}{\emph{CVPR}}. \bibinfo{pages}{1061--1069}.
\newblock


\bibitem[Xuey et~al\mbox{.}(2008)]%
        {xuey2008image}
\bibfield{author}{\bibinfo{person}{Su Xuey}, \bibinfo{person}{Jiaping Wang}, \bibinfo{person}{Xin Tong}, \bibinfo{person}{Qionghai Dai}, {and} \bibinfo{person}{Baining Guo}.} \bibinfo{year}{2008}\natexlab{}.
\newblock \showarticletitle{Image-based material weathering}. In \bibinfo{booktitle}{\emph{Comp. Graph. Forum}}, Vol.~\bibinfo{volume}{27}. \bibinfo{pages}{617--626}.
\newblock


\bibitem[Zhang et~al\mbox{.}(2020)]%
        {zhang2020dtvnet}
\bibfield{author}{\bibinfo{person}{Jiangning Zhang}, \bibinfo{person}{Chao Xu}, \bibinfo{person}{Liang Liu}, \bibinfo{person}{Mengmeng Wang}, \bibinfo{person}{Xia Wu}, \bibinfo{person}{Yong Liu}, {and} \bibinfo{person}{Yunliang Jiang}.} \bibinfo{year}{2020}\natexlab{}.
\newblock \showarticletitle{DTVNet: Dynamic Time-Lapse Video Generation via Single Still Image}. In \bibinfo{booktitle}{\emph{ECCV}}. \bibinfo{pages}{300--315}.
\newblock


\bibitem[Zhang et~al\mbox{.}(2021)]%
        {zhang2021dynamic}
\bibfield{author}{\bibinfo{person}{Kaitai Zhang}, \bibinfo{person}{Bin Wang}, \bibinfo{person}{Hong-Shuo Chen}, \bibinfo{person}{Ye Wang}, \bibinfo{person}{Shiyu Mou}, {and} \bibinfo{person}{C.-C.~Jay Kuo}.} \bibinfo{year}{2021}\natexlab{}.
\newblock \bibinfo{booktitle}{\emph{Dynamic Texture Synthesis by Incorporating Long-range Spatial and Temporal Correlations}}.
\newblock
\showeprint[arxiv]{2104.05940}~[cs]


\bibitem[Zhou et~al\mbox{.}(2023)]%
        {zhou2023photomat}
\bibfield{author}{\bibinfo{person}{Xilong Zhou}, \bibinfo{person}{Milos Hasan}, \bibinfo{person}{Valentin Deschaintre}, \bibinfo{person}{Paul Guerrero}, \bibinfo{person}{Yannick Hold-Geoffroy}, \bibinfo{person}{Kalyan Sunkavalli}, {and} \bibinfo{person}{Nima~Khademi Kalantari}.} \bibinfo{year}{2023}\natexlab{}.
\newblock \showarticletitle{Photomat: A material generator learned from single flash photos}. In \bibinfo{booktitle}{\emph{ACM SIGGRAPH}}. \bibinfo{pages}{1--11}.
\newblock


\bibitem[Zhou and Kalantari(2022)]%
        {zhou2022look}
\bibfield{author}{\bibinfo{person}{Xilong Zhou} {and} \bibinfo{person}{Nima~Khademi Kalantari}.} \bibinfo{year}{2022}\natexlab{}.
\newblock \showarticletitle{Look-ahead training with learned reflectance loss for single-image svbrdf estimation}.
\newblock \bibinfo{journal}{\emph{ACM Trans. Graph.}} \bibinfo{volume}{41}, \bibinfo{number}{6} (\bibinfo{year}{2022}), \bibinfo{pages}{1--12}.
\newblock


\end{thebibliography}



\begin{thebibliography}{10}


\ifx \showCODEN    \undefined \def \showCODEN     #1{\unskip}     \fi
\ifx \showDOI      \undefined \def \showDOI       #1{#1}\fi
\ifx \showISBNx    \undefined \def \showISBNx     #1{\unskip}     \fi
\ifx \showISBNxiii \undefined \def \showISBNxiii  #1{\unskip}     \fi
\ifx \showISSN     \undefined \def \showISSN      #1{\unskip}     \fi
\ifx \showLCCN     \undefined \def \showLCCN      #1{\unskip}     \fi
\ifx \shownote     \undefined \def \shownote      #1{#1}          \fi
\ifx \showarticletitle \undefined \def \showarticletitle #1{#1}   \fi
\ifx \showURL      \undefined \def \showURL       {\relax}        \fi
\providecommand\bibfield[2]{#2}
\providecommand\bibinfo[2]{#2}
\providecommand\natexlab[1]{#1}
\providecommand\showeprint[2][]{arXiv:#2}

\bibitem[Bradbury et~al\mbox{.}(2018)]%
        {bradbury2018jax}
\bibfield{author}{\bibinfo{person}{James Bradbury}, \bibinfo{person}{Roy Frostig}, \bibinfo{person}{Peter Hawkins}, \bibinfo{person}{Matthew~James Johnson}, \bibinfo{person}{Chris Leary}, \bibinfo{person}{Dougal Maclaurin}, \bibinfo{person}{George Necula}, \bibinfo{person}{Adam Paszke}, \bibinfo{person}{Jake VanderPlas}, \bibinfo{person}{Skye Wanderman-Milne}, {and} \bibinfo{person}{Qiao Zhang}.} \bibinfo{year}{2018}\natexlab{}.
\newblock \bibinfo{booktitle}{\emph{JAX: Composable Transformations of Python+NumPy Programs}}.
\newblock


\bibitem[Dupont et~al\mbox{.}(2019)]%
        {dupont2019augmented}
\bibfield{author}{\bibinfo{person}{Emilien Dupont}, \bibinfo{person}{Arnaud Doucet}, {and} \bibinfo{person}{Yee~Whye Teh}.} \bibinfo{year}{2019}\natexlab{}.
\newblock \showarticletitle{Augmented Neural ODEs}. In \bibinfo{booktitle}{\emph{NeurIPS}}, Vol.~\bibinfo{volume}{32}.
\newblock


\bibitem[Heitz(2014)]%
        {heitz2014understanding}
\bibfield{author}{\bibinfo{person}{Eric Heitz}.} \bibinfo{year}{2014}\natexlab{}.
\newblock \showarticletitle{Understanding the masking-shadowing function in microfacet-based BRDFs}.
\newblock \bibinfo{journal}{\emph{J Comp. Graph. Techniques}} \bibinfo{volume}{3}, \bibinfo{number}{2} (\bibinfo{year}{2014}), \bibinfo{pages}{32--91}.
\newblock


\bibitem[Kidger et~al\mbox{.}(2021)]%
        {kidger2021neural}
\bibfield{author}{\bibinfo{person}{Patrick Kidger}, \bibinfo{person}{James Foster}, \bibinfo{person}{Xuechen Li}, {and} \bibinfo{person}{Terry~J. Lyons}.} \bibinfo{year}{2021}\natexlab{}.
\newblock \showarticletitle{Neural SDEs as Infinite-Dimensional GANs}. In \bibinfo{booktitle}{\emph{ICML}}. \bibinfo{pages}{5453--5463}.
\newblock


\bibitem[Kidger and Garcia(2021)]%
        {kidger2021equinox}
\bibfield{author}{\bibinfo{person}{Patrick Kidger} {and} \bibinfo{person}{Cristian Garcia}.} \bibinfo{year}{2021}\natexlab{}.
\newblock \showarticletitle{Equinox: Neural Networks in {JAX} via Callable {PyTrees} and Filtered Transformations}.
\newblock \bibinfo{journal}{\emph{Differentiable Programming workshop at Neural Information Processing Systems 2021}} (\bibinfo{year}{2021}).
\newblock


\bibitem[Nikankin et~al\mbox{.}(2023)]%
        {nikankin2023sinfusion}
\bibfield{author}{\bibinfo{person}{Yaniv Nikankin}, \bibinfo{person}{Niv Haim}, {and} \bibinfo{person}{Michal Irani}.} \bibinfo{year}{2023}\natexlab{}.
\newblock \showarticletitle{SinFusion: training diffusion models on a single image or video}. In \bibinfo{booktitle}{\emph{ICML}}. Article \bibinfo{articleno}{1090}, \bibinfo{numpages}{16}~pages.
\newblock


\bibitem[Niklasson et~al\mbox{.}(2021)]%
        {niklasson2021self}
\bibfield{author}{\bibinfo{person}{Eyvind Niklasson}, \bibinfo{person}{Alexander Mordvintsev}, \bibinfo{person}{Ettore Randazzo}, {and} \bibinfo{person}{Michael Levin}.} \bibinfo{year}{2021}\natexlab{}.
\newblock \showarticletitle{Self-Organising Textures}.
\newblock \bibinfo{journal}{\emph{Distill}} \bibinfo{volume}{6}, \bibinfo{number}{2} (\bibinfo{year}{2021}), \bibinfo{pages}{10.23915/distill.00027.003}.
\newblock


\bibitem[Pajouheshgar et~al\mbox{.}(2023)]%
        {pajouheshgar2023dynca}
\bibfield{author}{\bibinfo{person}{Ehsan Pajouheshgar}, \bibinfo{person}{Yitao Xu}, \bibinfo{person}{Tong Zhang}, {and} \bibinfo{person}{Sabine S{\"u}sstrunk}.} \bibinfo{year}{2023}\natexlab{}.
\newblock \showarticletitle{Dynca: Real-time dynamic texture synthesis using neural cellular automata}. In \bibinfo{booktitle}{\emph{Proceedings of the IEEE/CVF Conference on Computer Vision and Pattern Recognition}}. \bibinfo{pages}{20742--20751}.
\newblock


\bibitem[Rombach et~al\mbox{.}(2022)]%
        {rombach2022high}
\bibfield{author}{\bibinfo{person}{Robin Rombach}, \bibinfo{person}{Andreas Blattmann}, \bibinfo{person}{Dominik Lorenz}, \bibinfo{person}{Patrick Esser}, {and} \bibinfo{person}{Björn Ommer}.} \bibinfo{year}{2022}\natexlab{}.
\newblock \showarticletitle{High-Resolution Image Synthesis with Latent Diffusion Models}. In \bibinfo{booktitle}{\emph{CVPR}}. \bibinfo{pages}{10674--10685}.
\newblock


\bibitem[Walter et~al\mbox{.}(2007)]%
        {walter2007microfacet}
\bibfield{author}{\bibinfo{person}{Bruce Walter}, \bibinfo{person}{Stephen~R Marschner}, \bibinfo{person}{Hongsong Li}, {and} \bibinfo{person}{Kenneth~E Torrance}.} \bibinfo{year}{2007}\natexlab{}.
\newblock \showarticletitle{Microfacet models for refraction through rough surfaces}. In \bibinfo{booktitle}{\emph{Proc. EGSR}}. \bibinfo{pages}{195--206}.
\newblock


\end{thebibliography}

\end{document}


\acmSubmissionID{281}
\setcopyright{acmlicensed}
\acmJournal{TOG}
\acmYear{2024} \acmVolume{43} \acmNumber{6} \acmArticle{256} \acmMonth{12}\acmDOI{10.1145/3687900}

\title{Neural Differential Appearance Equations: Supplemental Materials}

\author{Chen Liu}
\affiliation{%
	\institution{University College London}
	\country{United Kingdom}
}
\email{chen.liu.21@ucl.ac.uk}

\author{Tobias Ritschel}
\affiliation{%
	\institution{University College London}
	\country{United Kingdom}
}
\email{t.ritschel@ucl.ac.uk}

\maketitle

\mysection{Website}{website}

Besides this document, we provide two separate HTML files to showcase all our results and comparisons to baselines, for RGB and \ac{BRDF} results, respectively. Our project page is at \url{https://ryushinn.github.io/ode-appearance}.

\mysection{Implementation Details}{implementation}
Our implementation uses JAX \cite{bradbury2018jax} as the autodiff package with Equinox \cite{kidger2021equinox} and Diffrax \cite{kidger2021neural} as the neural \ac{ODE} solver package.
We conduct experiments in one Nvidia 4090 GPU with the following details.

\paragraph{Model.}
The model hyperparameters primarily include the depth of the UNet and the channel size at each level. 
For the appearance \acp{ODE} of RGB textures, we employ a three-level UNet with 32, 64, and 128 channels from top to bottom. 
The attention layers have 4 heads and 8 dimensions for each head. 
The time embedding has dimensions that are twice the channel size of the first level.
Appearance \acp{ODE} of \ac{BRDF} share the same backbone except being two-level with 64 and 128 at each, and not using attention layers. 

Our models are relatively lightweight, comprising 560K parameters for RGB \acp{ODE} and 500K parameters for \ac{BRDF} \acp{ODE}, in contrast to the millions of parameters typically found in modern diffusion models.

\paragraph{Training}
We train our RGB \acp{ODE} using Adam optimizer with 50,000 iterations and a $5\times 10^{-4}$ learning rate.
For \ac{BRDF} \acp{ODE} we train 60,000 iterations, with the initialization loss for the first 20,000 iterations and the crop loss for the rest.
For both tasks, we reduce the learning rate by half on plateaus of training loss, with the patience of 2500 iterations. 
We used $\text{batch-size}=1$ for all our experiments, which can already produces good results.
The average training time for each \ac{ODE} is 90 minutes.

We estimate Main Eq. 8 and Main Eq. 9 with 36 random crops and shuffles for each iteration.

We apply the common training techniques of neural \acp{ODE}, such as initializing \acp{ODE} to constant zero vector fields. 
During training, we also impose penalties on \ac{ODE} states that exceed their allowable range, such as negative RGB pixels or roughness values.

\paragraph{Time Range}
We heuristically choose $\timeWarmUp$, $\timeStart$, and $\timeEnd$ to match the temporal length of dynamics. 
For dynamic RGB, they are $-1$, $0$, and $5$ while they are scaled by two for dynamic \ac{BRDF}, as $-2$, $0$, and $10$.
We empirically observe that scaling does not influence the performance but the
ratio of warm-up duration ($[\timeWarmUp, \timeStart]$) to generation duration($[\timeStart, \timeEnd]$) is significant.
The ratio of 0.2 performs well for both tasks and every exemplar.

\paragraph{Training Algorithm}
To further validate our online training algorithm as in Main Alg. 1, we conduct experiments with the na\"ive estimate of Main. Eq. 5, \ie randomly sampling a time point in $[\timeStart, \timeEnd]$ to compute the loss for each iteration. 
For RGB \acp{ODE}, the average training time increases from 1.5 hours to 3.5 hours and the GRAM metric increases from 0.042 to 0.089.
Sometimes it even makes the training too unstable to finish.

We also experiment with different values of the refresh rate $\refreshRate$: 2, 4, 6, 8, and 10. 
The training times are 2.4, 1.7, 1.5, 1.4, and 1.2 hours, and the GRAM metrics are 0.071, 0.055, 0.042, 0.052, and 0.053, respectively. 
Hence we set $\refreshRate$ as 6 for the best balance.

\paragraph{Latent Space}
Our latent space of channel augmentation is suggested in \cite{dupont2019augmented} and previous works have revealed that the nine augmented channels are a good size \cite{niklasson2021self, pajouheshgar2023dynca}.

Nevertheless, we also do a sweeping of possible augmentation sizes for RGB \acp{ODE}: 0, 3, 6, 9, 15, and 21. 
They result in GRAM metrics of 0.127, 0.121, 0.048, 0.042, 0.049, and 0.058, and training time of 2.9, 1.7, 1.6, 1.5, 1.5, and 1.5 hours, respectively. 
So we choose 9 in the end.

We experiment with the latent space of a pretrained stable diffusion model \cite{rombach2022high} but the generated videos are notably less consistent. 
We conjecture that it is because the encoder and decoder are trained by images only, hence causing flickering artifacts even when decoding two close latent codes.
We believe that running our method on a pretrained latent space could be explored in future work.

\mysection{Training on temporally sparse data}{less_data}

\mycfigure{Lessdata}{
Our \ac{ODE} can be trained by only \textbf{F}ive \textbf{F}rames of the exemplar \material{Molding Bread}, which are framed in black in the first row.
As shown in the second row, the \ac{ODE} learns the essence of the underlying dynamics from these key-frames and faithfully reproduces the appearance in the unseen temporal regions.
}

\begin{table}[]
\caption{The numerical results for variants of our method. Ours (FF) is our method trained with only \textbf{F}ive \textbf{F}rames. Ours (\zeroOrder) is the zeroth order model. Ours (Iso.) is our \ac{svBRDF} \acp{ODE} with isotropic GGX distribution.}
\label{tab:suppl_tab}
\resizebox{\columnwidth}{!}{%
\begin{tabular}{llccccc}
\hline
 &  & \multicolumn{2}{c}{Gram $\downarrow$} & \multicolumn{2}{c}{SWD $\downarrow$} & Non-Str. $\downarrow$ \\ \hline
\multirow{3}{*}{\rotatebox[origin=c]{90}{RGB}} & Ours (\zeroOrder) & \multicolumn{2}{c}{0.057} & \multicolumn{2}{c}{0.064} & 0.720 \\
 & Ours (FF) & \multicolumn{2}{c}{0.107} & \multicolumn{2}{c}{0.101} & 0.677 \\
 & Ours & \multicolumn{2}{c}{0.042} & \multicolumn{2}{c}{0.050} & 0.720 \\ \hline
\multirow{4}{*}{\rotatebox[origin=c]{90}{\ac{BRDF}}} &  & \small{Center} & \small{Novel} & \small{Center} & \small{Novel} &  \\ \cline{3-6}
 & Ours (Iso.) & 0.021 & 0.020 & 0.036 & 0.040 & 0.984 \\
 & Ours (FF) & 0.049 & 0.038 & 0.069 & 0.063 & 0.986 \\
 & Ours & 0.020 & 0.020 & 0.034 & 0.038 & 0.984 \\ \hline
\end{tabular}%
}
\end{table}

In the main experiments, each input consists of 100 frames, sufficiently sampling the time domain from the start to the end of the dynamics.
However, samples might be lacking in the temporal space due to the difficulty of capturing continuously and stably. 
For instance, in an extreme case where only five key-frames are provided as input, it is evident that all our competitors would fail to produce any useful reproduction of the dynamic appearance as they primarily operate on a frame-by-frame basis.

In contrast, our method, benefiting from its inherent continuity of modeling the differential equations, can effectively interpolate between these under-sampled inputs, and even extrapolate to future times that have never been seen in the inputs.
One example is presented in \refFig{Lessdata}.

We also evaluate the performance of our method when supervised by limited data using the identical protocol of main experiments, as reported in \refTab{suppl_tab}. 
The results indicate that limited data primarily impacts the realism of the generated dynamic appearances. However, \acp{ODE} still achieve the same level of coherence, and even better coherence in the RGB case due to fewer constraints in the temporal domain.

\mysection{\firstOrder Order vs. \zeroOrder Order Dynamics}{1vs0}
Here we explore using a neural network to directly model the dynamic appearance itself (\zeroOrder order derivative).
The \zeroOrder order method should also be generative and operate on the temporal domain directly with inherent coherence.

SinFusion \cite{nikankin2023sinfusion}, which is conditioned on time to generate the frame directly from noise, is one example of \zeroOrder order modeling. 
But it substantially differs from our method in many aspects, such as goal, network design and size, training loss, etc.

We compare our \ac{ODE} method (modeling \firstOrder order derivative) to a \zeroOrder order method with a similar setting. Specifically, we train a network $
\ode_{\parameters}(\text{noise}, \timeCoord): 
(\stateRange^{\channels\times\height\times\width}, \timeRange)
\rightarrow 
\stateRange^{\channels\times\height\times\width},
$
to map a noise directly to the appearance at the given time $\timeCoord$.
We adopt the same network of our \ac{ODE} and the same loss for training. 

We experiment with this idea in our task of modeling dynamic RGB textures, referred to as ``Ours (\zeroOrder)'', since we are not aware of a publication.
The results are shown in \refTab{suppl_tab}. 
Ours (\zeroOrder) achieves the same coherence as our \acp{ODE}, while it is outperformed by our \acp{ODE} in terms of realism. 
We attribute the realism gap to the fact that learning a one-step denoising at each time point is more challenging than just modeling the difference between frames.
As analyzed in Main Sec. 6.3, the denoising (warm-up phase) is indeed a more difficult aspect of the dynamics. 
By using \acp{ODE}, we only need to model this part once.

\mysection{Anisotropy}{anisotropy}
\myfigure{aniso_iso}{Comparisons between results from our method with isotropic GGX (left), our method with anisotropic GGX (middle), and MatFusion (right), on one frame of an anisotropic material. We show two relit images for each method.}
We use an anisotropic GGX distribution \cite{heitz2014understanding} in our \ac{BRDF} model to account for the anisotropy effect in our real-world captures, prominently seen in the \material{Copper Patinating} sample.
In this section, we ablate this choice.

We train our \ac{svBRDF} \acp{ODE} with the isotropic GGX distribution \cite{walter2007microfacet} used commonly in our baselines and the literature, under the identical setting.
As shown in \refFig{aniso_iso}, a material that exhibits strong anisotropy, \eg the brushed copper film, fails to be modeled by the isotropic model, so as in our baselines, as anticipated.
This poses a challenge to approaches based on the supervision of (synthetic) ground-truth \ac{BRDF} maps, that the approach will struggle to reconstruct \acp{svBRDF} that fall out of the scope of \ac{BRDF} parameterization of training data.

The numerical results are shown in \refTab{suppl_tab}. The performance of these two variants is similar in both realism and coherence. Anisotropy seems not to impact the capability of modeling isotropic exemplars significantly.

\mysection{Dynamics Transfer}{dynamics_transfer}
\mycfigure{dynamics_transfer}{We train our \ac{ODE} using \material{Rusting Steel} data and apply it to three new instances: from left to right, a novel steel plate, a gold plate, and a silver plate.}

We can transfer the dynamics after learning the underlying \ac{ODE} of a dynamic appearance.
Our transfer does not involve re-training and is achieved by simulating the trained \ac{ODE} on a new image.

For example, after the rusting of a steel plate has been encoded into an \ac{ODE}, we can leverage it to produce a new trajectory of the rusting process. 
We showcase this with a new plate of steel, gold, and silver in \refFig{dynamics_transfer}.
In transfer, we do the same as novel dynamic texture synthesis except that we replace the synthesized image (\ie the first three channels of latent states) at $\timeStart$, when rusting begins, with our target image, and then simulate forward.

One issue is that the unchanged latent states will first transform our target image, such as a gold plate, back to a steel plate and then get it rusty.
This is not what we desire.
So we naively encode our target gold image to a steel-like image by channel-wise normalizing and then de-normalizing with the mean and deviation of the replaced states.
After obtaining the rusting process of the encoded steel plate, we decode it back to a gold plate using the reverse process.

Currently, our dynamics transfer can be only applied to semantically meaningful new images. That is, we can transfer rusting dynamics to a new metal plate, but not a slice of bread.
We believe that exploring more appropriate encoding functions is a promising research direction in the future, to migrate learned dynamics to a broader range of images.

\bibliographystyle{ACM-Reference-Format}
\bibliography{paper}